%% file: 0_main.tex
\documentclass[10pt]{article} % For LaTeX2e

\usepackage[accepted]{rlj} % Should be uncommented for the camera-ready
\usepackage{amssymb}            % Defines common symbols like \mathbb R
\usepackage{mathtools}          % Extends amsmath, providing common math tools
\usepackage{mathrsfs}           % Enables \mathscr, which can work in cases that \mathcal does not
\usepackage{graphicx}           % For including images
\usepackage{subcaption}         % Allows for the use of subfigures and subcaptions
\usepackage[space]{grffile}     % For spaces in image names
\usepackage{url}                % For displaying URLs
\usepackage{lipsum}             % For placeholder text
\usepackage{hyperref}

\usepackage{float}
\usepackage{amssymb}
\usepackage{amsmath}
\usepackage{booktabs}
\usepackage{subcaption}
\usepackage{graphicx}
\usepackage{enumitem}
\usepackage{xcolor}
\usepackage{multirow}
\usepackage{amsthm}
\usepackage{wrapfig}
\usepackage{etoc}
\usepackage[linesnumbered,ruled,vlined]{algorithm2e}
\usepackage{mwe}       % Provides example-image-A
\usepackage{titletoc}

\usepackage[table]{xcolor}
\definecolor{lightgray}{gray}{0.93}

\usepackage{booktabs, multirow, colortbl}
\newcommand{\taskwithimage}[2]{
    \item 
        \begin{minipage}[c]{0.15\linewidth}
            \centering
            \includegraphics[width=\linewidth]{#1}
        \end{minipage}
        \hfill
        \begin{minipage}[c]{0.8\linewidth}
        #2
        \end{minipage}
}

\newif\ifpreprint
\preprinttrue

\title{Simple Recipe Works: Vision-Language-Action \\Models are Natural Continual Learners with \\ Reinforcement Learning}

\setrunningtitle{Simple Recipe Works: Continual RL for VLA Models.}

\author{Jiaheng Hu\textsuperscript{1,$\ast$}, Jay Shim\textsuperscript{1,$\ast$}, Chen Tang\textsuperscript{2}, Yoonchang Sung\textsuperscript{3}, Bo Liu\textsuperscript{1}, \\
Peter Stone\textsuperscript{1,4,$\dagger$}, Roberto Mart\'{i}n-Mart\'{i}n\textsuperscript{1,$\dagger$}}

\emails{\{jiahengh,jshim1213,pstone,robertomm\}@utexas.edu}

\affiliations{
$^{1}$\textbf{UT Austin}  $^{2}$\textbf{UCLA}  $^{3}$\textbf{NTU}  $^{4}$\textbf{Sony AI}
\par % If including additional comments like below, use \par to add some whitespace. 
$\ast$ Indicating equal contribution  
$^\dagger$ Indicating equal supervision
}

\contribution{
    We conduct a comprehensive study of Continual Reinforcement Learning (CRL) for Vision-Language-Action (VLA) Models across diverse algorithms and benchmarks. In addition, we released our codebase to support reproducibility and future research.
    }
    {
    Existing CRL research has primarily focused on smaller models trained from scratch, and to our knowledge there has been no systematic study of continual RL in large Vision-Language-Action (VLA) models. Furthermore, there is currently no open-source framework designed for studying continual RL in this setting.
    }

\contribution{
    The results of our study lead to the surprising discovery that under the setup of parameter-efficient CRL for VLA models, naive \emph{Sequential Fine-Tuning} (Seq. FT) achieves strong performance, exhibiting high plasticity and minimal forgetting along with strong zero-shot generalization capabilities, and often outperforms the more complicated CRL methods. This finding establishes a simple but powerful baseline for CRL of VLA, and provide new insights into CRL for large pretrained models. 
    }
    {
    Conventional wisdom in the continual learning community suggests that naive Seq. FT leads to catastrophic forgetting. Our results show that under the parameter-efficient VLA fine-tuning paradigm, Seq. FT frequently outperforms sophisticated regularization and replay-based methods.
    }

\contribution{
    We provide a detailed analysis about the reason for the surprising effectiveness of Seq. FT. We found that the stability emerges from the synergy between the large pretrained model, parameter-efficient adaptation (LoRA), and on-policy RL, and justify this conclusion both empirically and from a mechanistic side.
    }
    {
    The stabilizing role of some of these components has been studied individually in prior work~\citep{shenfeld2025rlsrazoronlinereinforcement,biderman2024lora}. However, their combined effect in continual RL with large multimodal models has not previously been analyzed.
    }

 % Your keywords

\summary{
Continual Reinforcement Learning (CRL) for Vision-Language-Action (VLA) models is a promising direction toward self-improving embodied agents that can adapt in open-ended, evolving environments. However, conventional wisdom from continual learning suggests that naive Sequential Fine-Tuning (Seq. FT) leads to catastrophic forgetting, necessitating complex CRL strategies.
In this work, we take a step back and conduct a systematic study of CRL for large pretrained VLAs across three models and five challenging lifelong RL benchmarks. We find that, contrary to established belief, simple Seq. FT with low-rank adaptation (LoRA) is remarkably strong: it achieves high plasticity, exhibits little to no forgetting, and retains strong zero-shot generalization, frequently outperforming more sophisticated CRL methods.
Through detailed analysis, we show that this robustness arises from a synergy between large pretrained model, parameter-efficient adaptation, and on-policy RL. Together, these components reshape the stability–plasticity trade-off, making continual adaptation both stable and scalable. Our results position Sequential Fine-Tuning as a powerful method for continual RL with VLAs and provide new insights into lifelong learning in the large model era.
}

\begin{document}

\ifpreprint
% \makeCover  % Create the cover page
\else
\makeCover  % Create the cover page
\fi
\maketitle  % Make the title section

\begin{abstract}
Continual Reinforcement Learning (CRL) for Vision-Language-Action (VLA) models is a promising direction toward self-improving embodied agents that can adapt in open-ended, evolving environments. However, conventional wisdom from continual learning suggests that naive Sequential Fine-Tuning (Seq. FT) leads to catastrophic forgetting, necessitating complex CRL strategies.
In this work, we take a step back and conduct a systematic study of CRL for large pretrained VLAs across diverse lifelong RL benchmarks. We find that, contrary to established belief, simple Seq. FT with low-rank adaptation (LoRA) is remarkably strong: it achieves high plasticity, exhibits little to no forgetting, and retains strong zero-shot generalization, frequently outperforming more sophisticated CRL methods.
Through detailed analysis, we show that this robustness arises from a synergy between the large pretrained model, parameter-efficient adaptation, and on-policy RL. Together, these components reshape the stability–plasticity trade-off, making continual adaptation both stable and scalable. Our results position Sequential Fine-Tuning as a powerful method for continual RL with VLAs and provide new insights into lifelong learning in the large model era.\footnote{Code is available at \href{https://github.com/UT-Austin-RobIn/continual-vla-rl}{github.com/UT-Austin-RobIn/continual-vla-rl}.}
\end{abstract}

\input{1_intro}
\input{2_rw}

\input{3_prelim}

\input{4_1_experiment}

\input{4_2_add_exp}

\input{5_analysis}

%===============================================================================

\section{Conclusion}
\label{sec:conclusion}

In this work, we conducted a systematic study of Continual Reinforcement Learning for large Vision-Language-Action (VLA) models. 
Our investigation yielded a surprising and significant result: the simple approach of Sequential Fine-Tuning with Low-Rank Adaptation achieves strong plasticity, minimal forgetting, enhanced zero-shot generalization, and frequently outperforms more sophisticated CRL methods.
Further analysis reveals that this stability is not accidental but emerges from a synergy between the large pretrained model, parameter-efficient fine-tuning (LoRA), and the stable nature of on-policy RL post-training. These components collectively reshape the stability-plasticity dilemma, allowing the model to adapt to new tasks without overriding previous knowledge. 
Together, these findings offer us a simple but scalable recipe of how RL can be used as a powerful continual post-training paradigm for large pre-trained VLA models.

One natural future direction is to apply these findings to empower physical robotic systems, either via sim-to-real transfer~\citep{tobin2017domain,zhao2020sim} or real-world reinforcement learning~\citep{hu2025slac,zhu2020ingredients}. 
More generally, our results suggest that, as pre-trained models become larger and more capable, the traditional focus on catastrophic forgetting may no longer be the primary bottleneck in continual RL.
Instead, future work may benefit from designing algorithms that emphasize efficient adaptation and improved zero-shot generalization.
Ultimately, our findings and open-source codebase provide a principled starting point for the community to build more capable and adaptable lifelong embodied agents.

% Sanghavi, Sujay
\section*{Acknowledgments}
We thank Yifeng Zhu, Annie Xie, Sujay Sanghavi, Ben Abbatematteo, Zizhao Wang, Romir Sharma, and Kevin Rohling for their valuable feedback and discussions.
We thank members of LARG and UT Austin Machine Learning Laboratory for generously sharing computational resources that made this work possible, and the RLinf Team~\citep{yu2025rlinf} for the amazing infrastructure that this work built upon.
This work is supported in part by NSF
(FAIN-2019844, NRT-2125858), ONR (W911NF-25-1-0065), ARO
(W911NF-23-2-0004), Lockheed Martin, Amazon, and UT Austin's Good Systems grand
challenge. Jiaheng Hu is supported in part by a PhD fellowship from Two Sigma Investments, LP. 
Any opinions, findings, and conclusions or recommendations expressed in this material are those of the authors
and do not necessarily reflect the views of Two Sigma Investments. 
Peter Stone serves as the Chief Scientist of Sony AI and
receives financial compensation for that role.  The terms of this
arrangement have been reviewed and approved by the University of Texas
at Austin in accordance with its policy on objectivity in research.

%%%%%%%%%%%%%%%%%%%%%%%%%%%%%%%%%%%%%%%%%%%%%%%%%%%%%%%%%%%%%%%%
%% Appendices
%%%%%%%%%%%%%%%%%%%%%%%%%%%%%%%%%%%%%%%%%%%%%%%%%%%%%%%%%%%%%%%%

%%%%%%%%%%%%%%%%%%%%%%%%%%%%%%%%%%%%%%%%%%%%%%%%%%%%%%%%%%%%%%%%
%% NOTE: THIS MARKS THE END OF THE "MAIN TEXT"
%%%%%%%%%%%%%%%%%%%%%%%%%%%%%%%%%%%%%%%%%%%%%%%%%%%%%%%%%%%%%%%%

%%%%%%%%%%%%%%%%%%%%%%%%%%%%%%%%%%%%%%%%%%%%%%%%%%%%%%%%%%%%%%%%
%% Bibliography
%%%%%%%%%%%%%%%%%%%%%%%%%%%%%%%%%%%%%%%%%%%%%%%%%%%%%%%%%%%%%%%%
\bibliography{example}
\bibliographystyle{rlj}

%%%%%%%%%%%%%%%%%%%%%%%%%%%%%%%%%%%%%%%%%%%%%%%%%%%%%%%%%%%%%%%%
% AUTHOR: If your paper has no supplementary materials, you may 
%         comment out the line below, which creates the title for
%         the supplementary materials.
%%%%%%%%%%%%%%%%%%%%%%%%%%%%%%%%%%%%%%%%%%%%%%%%%%%%%%%%%%%%%%%%
\beginSupplementaryMaterials

\input{6_appendix}

\end{document}

%% file: 1_intro.tex
\section{Introduction}
\label{s:intro}

% VLA is good, but we need transform their initial generalization to lifelong competance
% To do this, we need to deal with the incremental improvement problem
Vision-Language-Action (VLA) models represent an emerging paradigm toward building general-purpose embodied agents. By fine-tuning VLMs for decision-making, these systems have demonstrated strong generalization across diverse scenarios~\citep{o2024open,kim2024openvla,black2024pi_0}. However, despite their broad competence, current VLA models remain brittle when deployed in evolving or out-of-distribution settings, where reliability and sustained adaptation become critical. This gap highlights the need for continual learning mechanisms that enable VLAs to incrementally refine and extend their capabilities through ongoing interaction, thereby transforming strong initial generalization into self-sustained, lifelong competence.

% CRL is a framework for doing that. CRL assumes non-stationarity. This is a great assumption but it also makes it very challenging: naive finetuning leads to catastrophic forgetting. Wile CRL methods constrain the update, they cause plasticity loss.
Such incremental self-improvement, where an agent needs to learn from a non-stationary stream of tasks and experiences, can be formalized as Continual Reinforcement Learning (CRL). 
The simplest approach to tackle CRL is through \emph{Sequential Fine-Tuning} (Seq. FT), where the model is directly finetuned on each new task or environments as it arrives. However, much prior work has shown that Seq. FT is prone to \textbf{catastrophic forgetting}, where the model’s performance on previously learned tasks degrades substantially as it adapts to new ones~\citep{french1999catastrophic,kirkpatrick2017overcoming,goodfellow2013empirical}.
To mitigate this effect, existing CRL methods introduce mechanisms such as regularization~\citep{kirkpatrick2017overcoming}, replay~\citep{rolnick2019experience,buzzega2020dark}, or parameter isolation~\citep{mallya2018packnet,yu2025moe} to constrain parameter updates. While these approaches are effective at preserving performance on previously learned tasks, they often come at the cost of \textbf{plasticity loss}, where the model’s ability to adapt to new tasks gradually diminishes. This trade-off between retaining past knowledge and remaining adaptable is known as the stability–plasticity dilemma, which poses a fundamental challenge for continual learning.

% Combining with VLA leads to even more challenges
The application to VLA models appears to make things even more difficult: on the one hand, modern VLA models contain billions of parameters and result in extremely computationally costly training. Therefore, efficient VLA post-training requires \emph{parameter-efficient fine-tuning} (PEFT) methods, such as LoRA~\citep{hu2021loralowrankadaptationlarge}, which in turn raises new questions about how PEFT interacts and potentially synergizes with CRL strategies.
On the other hand, these VLA models come with valuable pre-trained knowledge and strong zero-shot performance. As a result, we desire CRL algorithms that not only maintain the performance of trained tasks, but also preserve (and possibly enhance) these valuable \textbf{zero-shot generalization capabilities}.

% We hypothesize that continual learning behaves fundamentally differently in the regime of large pretrained VLA policies adapted with parameter-efficient updates and on-policy reinforcement learning. \textit{The combination of these three components naturally mitigates catastrophic forgetting, potentially rendering explicit continual learning mechanisms unnecessary}.

% We conduct a study. And naive sequential finetuning is enough!
How do existing CRL methods handle these aforementioned challenges? Does the interplay between large pretrained VLAs, PEFT adaptation, and RL introduce new technical difficulties? In this paper, we seek to answer these questions, by conducting a thorough empirical study of existing CRL methods across challenging lifelong RL benchmarks.
Our findings are striking. Across a wide range of CRL methods, the simple strategy of \textbf{Sequential Fine-Tuning} under standard low-rank adaptation (LoRA) consistently achieves \textbf{high plasticity and performances}, while exhibiting \textbf{little to no forgetting} and strong \textbf{zero-shot generalization} performance that often surpasses the multi-task oracle. 
In contrast, existing CRL methods, despite often making additional assumptions such as access to previous data and/or weights, consistently suffer from reduced plasticity due to their added constraints, leading to inferior adaptation to new tasks.

\begin{figure}[t]
\vspace{-1em}
\centering
% , trim=110 80 300 130, clip
\includegraphics[width=1.0\textwidth]{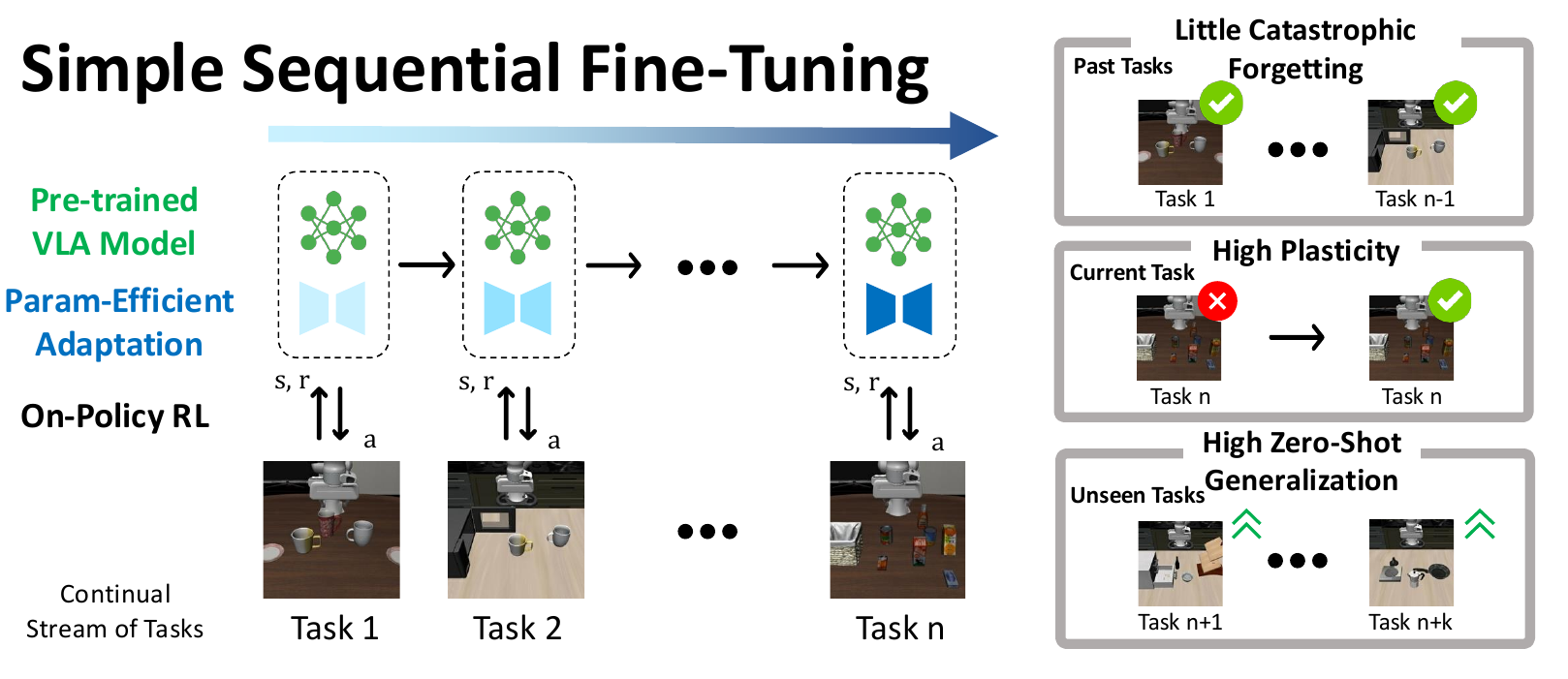}
\caption{
\textbf{VLAs as Natural Continual Learners}. We show that the synergy between pre-trained VLA, on-policy RL, and LoRA is enough to overcome catastrophic forgetting while maintaining plasticity, enabling simple Sequential Fine-Tuning to achieve surprisingly good performance.
}
\label{fig:env_viz}
\vspace{-2em}
\end{figure}

% What is the implication? Why does this happen? Our analysis show that...
These findings are exciting because they reveal an unexpectedly simple yet highly effective path toward scalable lifelong adaptation in large VLAs. However, they are also quite puzzling, since they stand in stark contrast to previous results from the continual learning community, where Sequential Fine-Tuning typically leads to severe forgetting and thus low performance.
Upon further investigation, we find that the robustness of naive finetuning emerges from the interplay between large pre-trained VLAs, LoRA-based parameter-efficient adaptation, and on-policy reinforcement learning. Rather than exacerbating instability, these components collectively make continual adaptation more stable, while synergistically preserving the learning plasticity. More specifically, our analysis finds that each of these three components mitigates catastrophic forgetting from a complementary perspective, and removing any single one of them causes a significant increase in forgetting.
Taken together, our results and analysis establish parameter-efficient Sequential Fine-Tuning as a simple but effective method for continual reinforcement learning with VLA models. These results, supported by our open-source implementation, offer a principled starting point for future work on scalable lifelong embodied intelligence.

% \begin{figure}[t]
% \centering
%     \begin{subfigure}[b]{0.24\textwidth}
%         \centering
%         \includegraphics[width=\linewidth]{images/wipe_env_sim_resize.png}
%         \caption{Simulation Board}
%         \label{fig:sb}
%     \end{subfigure}
%     \hfill
%     \begin{subfigure}[b]{0.24\textwidth}
%         \includegraphics[width=\linewidth]{images/board_resize_2.jpg}
%         \caption{Real-world Board}
%         \label{fig:rb}
%     \end{subfigure}
%     \hfill
%     % \medskip
%     % \vspace{-1\baselineskip}
%     \begin{subfigure}[b]{0.24\textwidth}
%         \centering
%         \includegraphics[width=\linewidth]{images/table_env_sim_resize.png}
%         \caption{Simulation Table}
%         \label{fig:st}
%     \end{subfigure}
%     \hfill
%     \begin{subfigure}[b]{0.24\textwidth}
%         \includegraphics[width=\linewidth]{images/table_real_resize_2.jpg}
%         \caption{Real-world Table}
%         \label{fig:rt}
%     \end{subfigure}
%     \hfill
% \caption{\methodname{} use action space trained in low-fidelity simulation to learn downstream tasks in the real world. It learns to solve challenging contact-rich bimanual whole-body mobile manipulation tasks with less than an hour of real-world interactions.
% }
% \label{fig:env_viz}
% \end{figure}

%% file: 2_rw.tex
\section{Background \& Related Work}

\label{sec:rw}

\paragraph{Vision-Language-Action Models.} VLA models unify visual perception, natural-language conditioning, and action generation in a single policy. They are typically trained on large-scale robot datasets by imitation learning which results in generalization capability across tasks and environments. A major family of models adopts \emph{autoregressive} action generation: RT-1, RT-2, and OpenVLA~\citep{Brohan2022RT1RT,Brohan2023RT2VM,Kim2024OpenVLAAO} discretize actions into tokens and decode them auto-regressively conditioned on images and task instructions. A closely related variant uses \emph{action chunking}, where the
policy predicts short action horizons at each decision step rather
than a single action, with OpenVLA-OFT as a representative example~\citep{kim2025finetuningvisionlanguageactionmodelsoptimizing}. Another family of approaches uses \emph{continuous generative} action heads: diffusion-based
policies generate actions through iterative denoising~\citep{Chi2023DiffusionPV}, while Pi-0 adopts a flow-matching head built on a vision-language
backbone as an alternative continuous-action VLA design~\citep{black2024pi_0}.

\paragraph{Reinforcement Learning Post-Training of VLA Models.}
RL post-training recently emerged as an effective methodology to refine and improve large pretrained Vision-Language-Action (VLA) models~\citep{deng2025survey,lu2025vla,hu2024flare,yu2025rlinf,intelligence2025pi,chen2026pitextttrlonlinerlfinetuning,wagenmaker2025steering}.
The pretrained generalization capabilities of VLAs allow for effective exploration and open up exciting possibilities for learning from sparse rewards on challenging tasks. 
A key challenge in RL post-training of VLA Models is maintaining training stability and avoiding performance collapse. 
Prior work has shown that stable adaptation requires carefully controlled on-policy updates, small learning rates, and well-behaved policy objectives~\citep{hu2024flare,yu2025rlinf}. 

Following this established recipe, we adopt on-policy reinforcement learning throughout this work. 
In particular, we use Group Relative Policy Optimization (GRPO)~\citep{guo2025deepseek}, a stable policy-gradient method that has achieved strong empirical performance in large-scale post-training.
We provide a detailed description of GRPO and its application for training autoregressive and flow-based VLAs in Appendix.~\ref{app:grpo}.

\paragraph{Continual Reinforcement Learning.}
Continual reinforcement learning (CRL)~\citep{pan2025survey,abbas2023loss,dohare2024loss,tang2025mitigatingplasticitylosscontinual,khetarpal2022towards,meng2025preserving,mesbahi2025positionlifetimetuningincompatible,abel2023definition,elelimy2025rethinking} studies RL agents that must adapt continually to non-stationary tasks or environments while retaining competence on previously encountered ones. A common categorization is \emph{what} is transferred across changes~\citep{wolczyk2022disentangling, pan2025survey}, such as value functions~\citep{anand2023predictioncontrolcontinualreinforcement}, policies~\citep{kaplanis2019policy,berseth2021comps}, experiences~\citep{xie2022lifelong}, or learned dynamics models~\citep{kessler2023effectiveness}, and \emph{how} transfer is implemented~\citep{pan2025survey,khetarpal2022towards}, which can be grouped into: (i) \emph{regularization-based} methods that constrain parameter updates to reduce interference~\citep{kirkpatrick2017overcoming}, (ii) \emph{replay-based} methods that preserve and reuse past experience~\citep{rolnick2019experience,buzzega2020dark}, and (iii) \emph{parameter-isolation} methods that allocate additional state or parameters to isolate or store knowledge~\citep{rusu2016progressive}. Most of these works only consider small models trained from scratch. By contrast, we focus on CRL applied to large pre-trained VLA models and the intriguing properties that arise from such a setup.

% In this work, we focus on \emph{policy}-based CRL because our target setting is continual post-training of a pre-trained VLA policy that is not necessarily paired with a value function or learned dynamics model from pre-training. 

% \jiaheng{Merge with Preliminaries}
% \paragraph{Continual Learning on Large Models.} Scaling to large pretrained backbones changes the regime of continual learning: full fine-tuning is often prohibitively expensive and can exacerbate catastrophic forgetting. Parameter-efficient fine-tuning (PEFT), especially LoRA, therefore becomes the natural adaptation tool, enabling strong transfer with a small number of trainable parameters~\citep{hu2021loralowrankadaptationlarge}. Recent work extends PEFT to \emph{parameter-efficient} continual learning under tight memory and compute budgets, for example, through the combination of low-rank transfer objectives and gradient-projection techniques~\citep{Qiao2024LearnMB}, continual merging methods that avoid accumulating per-task adapters~\citep{Qiao2025MergeBF,Marczak2024MagMaxLM}, and dynamic rank allocation over time to balance plasticity and retention~\citep{Bhat2025ParameterEC}. Most of these studies are conducted in supervised or offline continual learning regimes (often on LLMs); in contrast, our focus is \emph{online, on-policy} continual adaptation for embodied control, where distribution shift is induced by both environment changes and the agent's evolving policy. 

% Citations to add:~\citep{romer2026clarecontinuallearningvisionlanguageaction} 
% \citep{zheng2023preventing} 
% \citep{shi2025continual}

\paragraph{Parameter-Efficient Fine-Tuning.}

Given the scale of modern generative models such as VLAs, full-parameter fine-tuning is often prohibitively expensive, especially in continual learning settings~\citep{shi2025continual}. 
This has motivated parameter-efficient fine-tuning (PEFT)~\citep{fu2023effectiveness,ding2023parameter,li2021prefix,hu2021loralowrankadaptationlarge}, which adapts a pretrained network by updating only a small subset of parameters while keeping the backbone weights frozen. Among various PEFT methods, the predominant approach is Low-Rank Adaptation (LoRA)~\citep{hu2021loralowrankadaptationlarge, liu2023tail,Qiao2024LearnMB}. 
LoRA adapts a pretrained model by parameterizing weight updates as low-rank matrices while keeping the original pretrained weights frozen.
Concretely, for a pretrained weight matrix $W_0 \in \mathbb{R}^{d \times k}$, LoRA parametrizes the adapted weight as
\[
W = W_0 + BA,
\]
where $B \in \mathbb{R}^{d \times r}$ and $A \in \mathbb{R}^{r \times k}$ are trainable matrices with rank $r \ll \min(d,k)$. After training, the LoRA weight can be easily merged into the original weight via $W_\text{new} \leftarrow W_0 + BA$.

This formulation significantly reduces the number of trainable parameters while preserving the expressive capacity of the pretrained model. 
Given its strong empirical performance and widespread adoption in large-scale model adaptation, we adopt LoRA as our parameter-efficient fine-tuning method throughout this work.

% \paragraph{Action chunking and diffusion/flow-based action heads.}
% Beyond backbone and data, VLA systems differ substantially in their \emph{action representation} and decoding procedure.
% Action chunking---predicting a short horizon of actions per decision---is a common technique to improve temporal consistency and reduce per-step planning overhead, but can introduce additional decoding cost and complicate continual adaptation objectives.
% Recent work explores accelerating chunked VLA inference through parallel decoding~\citep{song2025accelerating}.
% Another strong design choice is the use of generative action heads.
% Diffusion-based policies model action trajectories as a denoising process and have been shown to be highly effective for visuomotor control~\citep{Chi2023DiffusionPV}.
% More recently, flow-based formulations have been proposed for general robot control in VLA-style settings~\citep{black2024pi_0}.
% Since our goal is continual adaptation of a pre-trained VLA policy, these architectural choices (chunking and diffusion/flow heads) directly influence both the optimization dynamics and the stability--plasticity trade-off we observe during online continual learning.

%% file: 3_prelim.tex
\section{Problem Formulation}

\subsection{Language-Conditioned MDP for VLA Post-Training}
We formulate each task in VLA post-training as a finite-horizon, language-conditioned Markov Decision Process (MDP):
\[
\mathcal{M} = (\mathcal{S}, \mathcal{A}, P, H, \mu_0, \ell, r),
\]
where $\mathcal{S}$ denotes the state space, 
$\mathcal{A}$ denotes the action space, 
$P : \mathcal{S} \times \mathcal{A} \rightarrow \mathcal{S}$ is the transition function, 
$H$ is the horizon, 
$\mu_0$ is the initial state distribution, 
% $\gamma \in (0, 1]$ is the discount factor
$\ell \in \mathcal{L}$ is a natural-language instruction specifying the task,
and $r : \mathcal{S} \times \mathcal{A} \times \mathcal{L} \rightarrow \{0,1\}$ is a \textbf{sparse reward function}. For each task, the VLA policy $\pi_\theta(a_t \mid s_t, \ell)$ is trained to maximize the cumulative reward.

In our work, all tasks share the same state and action space, where the state space consists of camera images, and the action space consists of robot end-effector pose and gripper command.

\subsection{Continual Reinforcement Learning in Language-Conditioned MDPs}

In the continual setting, the agent learns sequentially over $T$ tasks\footnote{We note that, while some prior continual reinforcement learning formulations assume the task identity is latent or unobserved~\citep{khetarpal2022towards}, 
in our setting the task specification is directly provided as natural language input to the VLA model. Since the policy is explicitly conditioned on $\ell$, 
it is both natural and necessary to assume that the task instruction is observable to the agent.
}
$\{\mathcal{T}_1, \dots, \mathcal{T}_T\}$ in fixed order that is beyond the control of the agent, where each task $\mathcal{T}^k$ is represented by a language instruction $\ell^k$ and its corresponding sparse reward function $r^k$. Up to task $k$, the CRL objective is to optimize the average return over all seen tasks:
\[
\max_\theta
J_{\mathrm{CRL}}(\theta)
=
\frac{1}{k}
\sum_{j=1}^{k}
\mathbb{E}_{\pi_\theta}
\left[
\sum_{t=1}^{H} r^j
\right].
\]
The agent learns each task purely through interacting with the environment, \textbf{without access to any demonstrations}.
A defining characteristic of the CRL setting is that, when learning task $\mathcal{T}^k$, 
the agent cannot access data or interact with the environments of previous tasks 
$\{\mathcal{T}^1, \dots, \mathcal{T}^{k-1}\}$.

\subsection{Evaluation Metrics}
\label{ss:metrics}
Following the existing literature~\citep{lopez2017gradient,chaudhry2019tiny,zheng2023preventing,abel2023definition}, we adopt standard continual learning metrics for performance evaluation, including \emph{Average Success (AVG)}, which measures overall performance at the end of training, \emph{Negative Backward Transfer (NBT)}, which measures forgetting, and \emph{Forward Transfer (FWT)}, which measures generalization. In addition, we introduce \emph{Zero-Shot Success (ZS)} as a new metric to measure the ability of the algorithm to retain pre-trained capabilities in the VLA. We describe these metrics in detail in the supplementary material (Appendix~\ref{app:ev_metric}).

%% file: 4_1_experiment.tex
\begin{figure}[t]
    \centering
    % ---- Left figure with its own caption ----
    \begin{subfigure}[t]{0.144\textwidth}
        \centering
        \includegraphics[width=\textwidth]{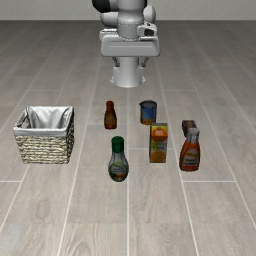}
        \caption{LB-Object}
    \end{subfigure}
    % ---- Left figure with its own caption ----
    \begin{subfigure}[t]{0.144\textwidth}
        \centering
        \includegraphics[width=\textwidth]{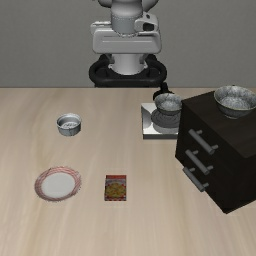}
        \caption{LB-Spatial}
    \end{subfigure}
    \begin{subfigure}[t]{0.144\textwidth}
        \centering
        \includegraphics[width=\textwidth]{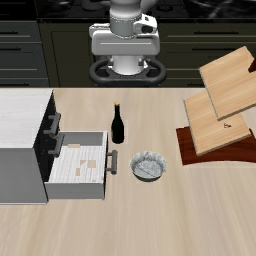}
        \caption{LB-Long}
    \end{subfigure}
    \begin{subfigure}[t]{0.156\textwidth}
        \centering
        \includegraphics[width=\textwidth, trim=15 25 15 20, clip]{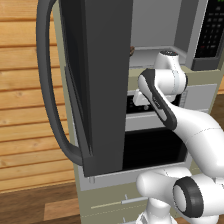}
        \caption{RoboCasa}
    \end{subfigure}
    \begin{subfigure}[t]{0.192\textwidth}
        \centering
        \includegraphics[width=\textwidth]{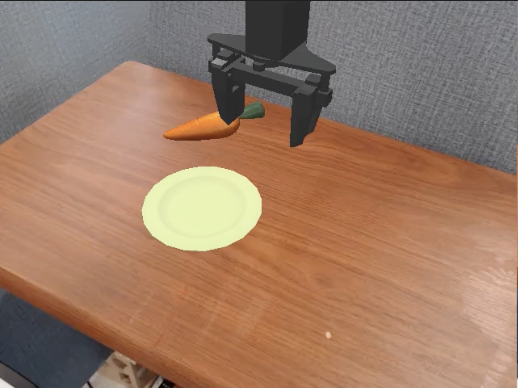}
        \caption{Maniskill}
    \end{subfigure}

    \caption{Our evaluation spans diverse tasks and benchmarks. Here we show one task from each benchmark. For visualization and description of all the tasks, see Appendix~\ref{app:env_desc}.}
    \label{fig:full_row}
    \vspace{-2em}
\end{figure}

\section{An Empirical Study of Continual RL for VLAs}
\label{sec:exp}

In this section, we empirically evaluate continual reinforcement learning (CRL) methods for post-training large Vision-Language-Action (VLA) models.
In Sec.~\ref{ss:setup}, we describe the experiment setup and algorithms. In Sec.~\ref{ss:main_rslt} and Sec.~\ref{ss:perturb}, we present our results and findings. 
% Finally, in Sec.~\ref{ss:additional}, we present additional results and discussions based on the findings from Sec.~\ref{ss:main_rslt} and Sec.~\ref{ss:perturb}.
Finally, in Sec.~\ref{ss:additional}, we present additional results and discussions regarding closing the training gap with oracle (Sec.~\ref{ss:gap}), the properties of VLAs with continuous diffusion head (Sec.~\ref{sss:diffusion}), and increasing the length of training task sequence (Sec.~\ref{sss:long}).

% Our primary goal is to understand whether specialized CRL techniques are necessary in this regime, or whether simple sequential fine-tuning is sufficient. 
% We evaluate methods across multiple VLA backbones and lifelong RL benchmarks, with a focus on catastrophic forgetting, plasticity, and zero-shot generalization.

\subsection{Experimental Setup}
\label{ss:setup}

We follow a consistent training protocol across all methods to ensure fair comparison. 
As explained in Sec.~\ref{sec:rw}, all of our experiments are conducted with GRPO and LoRA unless noted otherwise.
Specifically, all methods share the same core hyperparameters, including network architecture, learning rate, batch size, optimizer config, LoRA rank, and GRPO hyperparameters, which we directly inherit from the default configuration of~\citet{yu2025rlinf}. For method-specific hyperparameters (e.g., EWC coefficient, Replay coefficient), we perform a local sweep within one order of magnitude of the values reported in the original papers and select the best-performing setting.  
Notably, we do not do any hyperparameter tuning for \emph{Sequential Fine-Tuning}.
We provide additional details in the supplementary material, including details regarding the base VLA, pretraining datasets, train/heldout splits, and training durations (Appendix~\ref{app:setup}), as well as shared and method-specific hyperparameters (Appendix~\ref{app:shared_hype}-\ref{app:sepc_hype}).
\ifpreprint
We aggregate results across 3 independent random seeds for each experiment and report mean $\pm$ standard error. 
\else

The high computation cost of fine-tuning large VLA models constrains the number of random seeds we can run per experiment. 
Therefore, following established protocols in the Libero benchmark~\citep{liu2023libero}, we aggregate results across 3 independent random seeds for each experiment and report mean $\pm$ standard error. 
Our experiments took a total of approximately \emph{3600 GPU-days} to run on NVIDIA A100 GPUs.
Given this constraint, our analysis prioritizes identifying consistent trends across a diverse suite of benchmarks rather than claiming absolute statistical significance.
\fi
% \jiaheng{The best performance is bolded, and colored in purple if the improvement is statistically significant over other algorithms, when a two-tailed, Student’s t-test under equal sample sizes and unequal variance is applied with a p-value of 0.05.}

\paragraph{CRL Algorithms}
We focus our evaluation on eight algorithms spanning the dominant paradigms in Continual Reinforcement Learning. As reference points, Sequential Fine-Tuning (often used in prior work as lower bound) trains tasks sequentially without any forgetting-prevention mechanism, while Multi-Task Training (upper bound oracle) breaks the non-stationary assumption and trains jointly on all tasks simultaneously. Next, we evaluate representatives of the three principal CRL paradigms~\citep{pan2025survey}: Elastic Weight Consolidation~\citep{kirkpatrick2017overcoming} (regularization-based), Expert Replay~\citep{rolnick2019experience} and Dark Experience Replay~\citep{buzzega2020dark} (replay-based), and Dynamic Weight Expansion (parameter isolation~\citep{rusu2016progressive}). We additionally evaluate two methods motivated by large pretrained model adaptation: SLCA~\citep{zhang2023slca}, which applies layerwise learning-rate decoupling to preserve pretrained representations, and RETAIN~\citep{yadav2025robust}, which uses discounted weight merging to balance adaptation and retention. Full descriptions of each algorithm are provided in Appendix~\ref{app:algo}.

\subsection{Results: A Study of CRL Methods on VLAs}
\label{ss:main_rslt}
\paragraph{Evaluation Domains}

For the first set of experiments, we evaluate on three benchmarks: libero-object, libero-spatial, and libero-long-horizon. All three benchmarks consist of challenging robot manipulation tasks, with each focusing on different aspects of knowledge transfer\footnote{Note that while some recent papers claimed high success rate on the ``libero benchmarks'', they are often ignoring the continual learning assumptions, training on the test tasks, training without considering the epoch limits, and/or training with expert demonstrations, which makes those results inapplicable to our problem setup.}. Although the LIBERO benchmarks provide expert demonstrations, we do not use these demonstrations during continual post-training, except in the ER method, where they are used for replay. In each of these tasks, the VLA model takes in an RGB image and a natural-language instruction, and outputs a sequence of 7-dimensional actions that controls the end-effector poses and the gripper state. We visualize these benchmarks in Fig.~\ref{fig:full_row}, and refer the reader to ~\cite{liu2023libero} for a more detailed description of these tasks. We present the results in Table~\ref{tab:main}. 

\begin{table}[h]
\vspace{-1.0em}
\caption{Comparison of performance across CRL algorithms. Each number represent success rate of tasks (\%). In addition to the metrics discussed in Sec.~\ref{ss:metrics}, we report $\Delta$ between the initial checkpoint and the final checkpoint to indicate performance change during training. We \textbf{bold} the highest-performing method for each metric, not including the multitask oracle.}
\centering
\renewcommand{\arraystretch}{1.3}
\footnotesize
\begin{tabular}{lcccccc}
\toprule
\textbf{Domain} / Method & \multicolumn{6}{c}{\textbf{Metrics (\%)}} \\
\cmidrule(lr){2-7}
& \textbf{AVG} $\uparrow$ & \textbf{$\Delta$AVG} $\uparrow$ & \textbf{NBT} $\downarrow$ & \textbf{FWT} $\uparrow$ & \textbf{ZS} $\uparrow$ & \textbf{$\Delta$ZS} $\uparrow$ \\
\midrule

\multicolumn{7}{l}{\textbf{libero-spatial}} \\
\rowcolor{blue!10}\quad Sequential Fine-Tuning & \textbf{81.2$\pm$0.4} & \textbf{+24.3} & 0.3$\pm$0.5 & \textbf{3.9$\pm$1.5} & \textbf{57.1$\pm$1.1} & \textbf{+5.6} \\
\quad Elastic Weight Consolidation      & 66.1$\pm$0.9 & +9.3 & 0.7$\pm$1.7 & 1.5$\pm$0.3 & 52.6$\pm$0.9 & +1.1 \\
\quad Expert Replay       & 80.2$\pm$0.5 & +23.3 & 0.6$\pm$1.1 & -2.3$\pm$0.1 & 49.2$\pm$1.0 & -2.3 \\
\quad Dark Experience Replay      & 73.4$\pm$1.3 & +16.6 & 4.7$\pm$1.3 & 0.7$\pm$0.9 & 55.2$\pm$0.7 & +3.7 \\
\quad Dynamic Weight Expansion      & 79.6$\pm$0.9 & +22.7 & 0.0$\pm$0.0 & 0.0$\pm$0.0  & 51.5$\pm$0.0 & +0.0 \\
\quad SLCA (Layered LR)     & 69.9$\pm$0.7 & +13.0 & \textbf{-0.6$\pm$2.0} & 1.5$\pm$0.3 & 56.1$\pm$0.9 & +4.6 \\
\quad RETAIN (Weight Merging)   & 66.0$\pm$0.7 & +9.1 & 2.9$\pm$1.4 & 1.4$\pm$1.4 & 53.7$\pm$0.8 & +2.2 \\
\rowcolor{gray!10} \quad Multitask (Oracle) & 85.8$\pm$0.2 & +28.9 & -- & -- & 51.2$\pm$0.7 & -0.3 \\

\midrule

\multicolumn{7}{l}{\textbf{libero-object}} \\
\rowcolor{blue!10}\quad Sequential Fine-Tuning & \textbf{93.2$\pm$0.7} & \textbf{+37.6} & 1.0$\pm$0.7 & 7.1$\pm$0.8 & 25.4$\pm$0.2 & +5.8 \\
\quad Elastic Weight Consolidation      & 82.6$\pm$1.2 & +26.9 & 0.1$\pm$0.8 & \textbf{10.0$\pm$0.4} & 25.3$\pm$0.8 & +5.6 \\
\quad Expert Replay       & 88.8$\pm$0.2 & +33.1 & 4.5$\pm$0.6 & 6.4$\pm$1.1 & \textbf{26.7$\pm$0.5} & \textbf{+7.1} \\
\quad Dark Experience Replay      & 89.1$\pm$0.2 & +33.4 & 0.8$\pm$1.1 & 6.8$\pm$0.8 & 24.8$\pm$1.7 & +5.2 \\
\quad Dynamic Weight Expansion      & 92.4$\pm$0.3 & +36.7 & 0.0$\pm$0.0 & 0.0$\pm$0.0 & 19.6$\pm$0.0 & +0.0 \\
\quad SLCA (Layered LR)     & 84.1$\pm$0.7 & +28.4 & \textbf{-1.6$\pm$0.5} & +5.2$\pm$1.4 & 24.2$\pm$0.2 & +4.6 \\
\quad RETAIN (Weight Merging)   & 76.6$\pm$0.3 & +20.9 & 0.8$\pm$1.0 & 1.8$\pm$1.5 & 22.5$\pm$0.9 & +2.9 \\
\rowcolor{gray!10} \quad Multitask (Oracle) & 95.7$\pm$0.7 & +40.1 & -- & -- & 27.6$\pm$1.3 & +8.0 \\

\midrule

\multicolumn{7}{l}{\textbf{libero-long-horizon}} \\
\rowcolor{blue!10}\quad Sequential Fine-Tuning & \textbf{89.8$\pm$0.9} & \textbf{+6.8} & \textbf{-2.4$\pm$1.0} & 0.5$\pm$0.1 & 86.6$\pm$0.2 & +3.3 \\
\quad Elastic Weight Consolidation      & 86.6$\pm$0.3 & +3.6 & 0.8$\pm$1.3 & \textbf{3.0$\pm$1.3} & 86.5$\pm$0.1 & +3.1 \\
\quad Expert Replay       & 88.8$\pm$0.8 & +5.8 & -0.2$\pm$1.7 & -1.1$\pm$0.6 & 83.2$\pm$0.2 & -0.1 \\
\quad Dark Experience Replay      & 87.6$\pm$0.4 & +4.6 & 0.7$\pm$0.8 & 0.7$\pm$0.2 & 84.7$\pm$0.2 & +1.3 \\
\quad Dynamic Weight Expansion      & 88.4$\pm$0.5 & +5.4 & 0.0$\pm$0.0 & 0.0$\pm$0.0 & 83.4$\pm$0.0 & +0.0 \\
\quad SLCA (Layered LR)     & 86.9$\pm$0.6 & +3.9 & -1.3$\pm$1.0 & -0.2$\pm$0.3 & 86.1$\pm$0.7 & +2.7 \\
\quad RETAIN (Weight Merging)   & 86.2$\pm$0.9 & +3.2 & 1.6$\pm$1.0 & 1.0$\pm$1.2 & \textbf{86.9$\pm$0.2} & \textbf{+3.6} \\
\rowcolor{gray!10} \quad Multitask (Oracle) & 90.5$\pm$0.8 & +7.5 & -- & -- & 85.2$\pm$0.5 & +1.8 \\

\bottomrule
\end{tabular}
\vspace{-1em}
\label{tab:main}
\end{table}

\begin{wrapfigure}{r}{0.5\textwidth}
    \vspace{-1.5em}
    \centering
    % ---- Left figure with its own caption ----
    \begin{subfigure}[t]{0.25\textwidth}
        \centering
        \includegraphics[width=\textwidth]{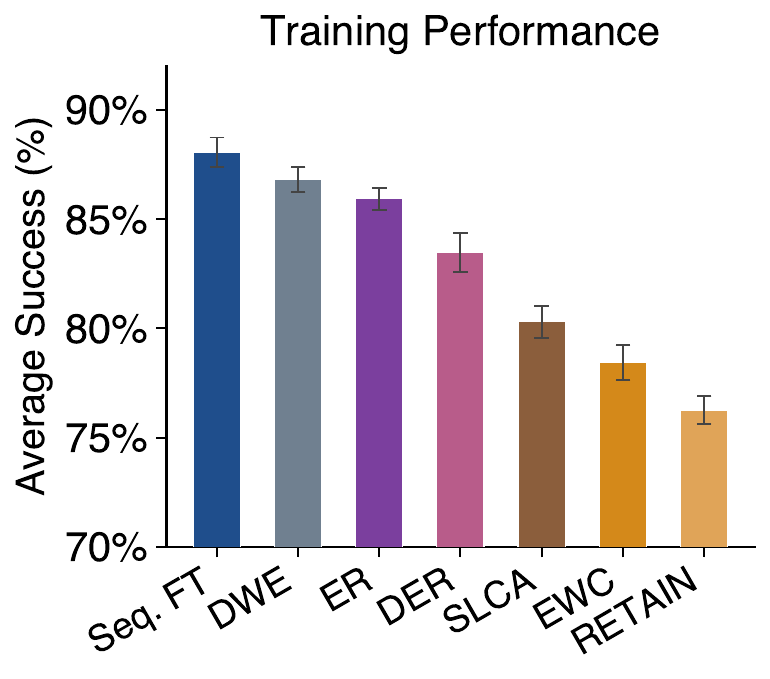}
    \end{subfigure}\hfill
    % ---- Left figure with its own caption ----
    \begin{subfigure}[t]{0.25\textwidth}
        \centering
        \includegraphics[width=\textwidth]{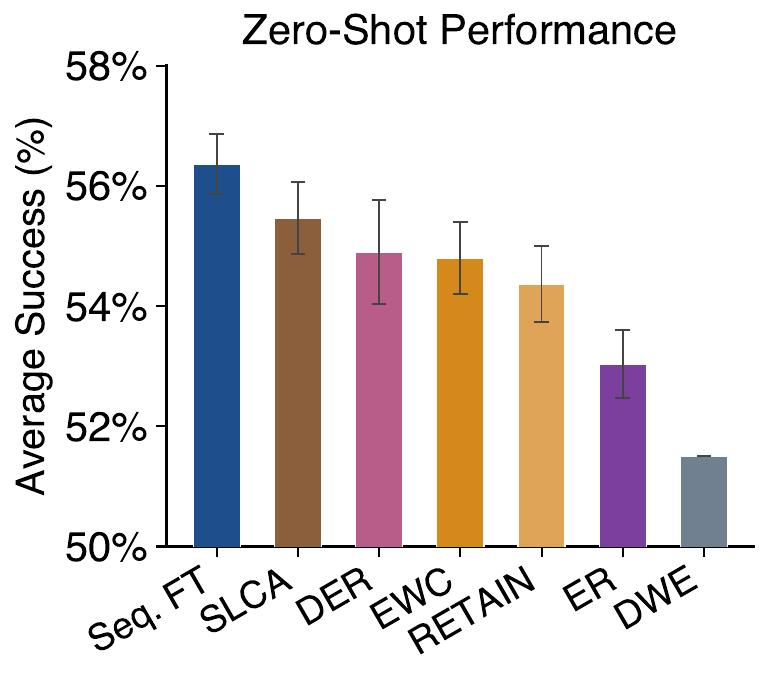}
    \end{subfigure}\hfill
    \vspace{-1em}
    \caption{Averaged across three benchmarks, Seq, FT obtains strong performance in both performance (AVG) and generalization (ZS).}
    \label{fig:performance}
    \vspace{-1em}
\end{wrapfigure}

Across the three benchmarks, Sequential Fine-Tuning (Seq. FT) consistently achieves strong performance (Fig.~\ref{fig:performance}).
In terms of \textbf{Average Success on the training tasks (AVG)}, Seq. FT achieves performance similar to replay-based and parameter isolation methods, and surpasses the rest of the CRL methods.
While the average training success of Seq. FT is often slightly lower than the multi-task oracle, this gap is generally quite small and can be closed under modest modifications to the training setup, as we will demonstrate in Sec.~\ref{ss:gap}.

In the meantime, Sequential Fine-Tuning consistently preserves \textbf{strong zero-shot generalization capabilities},
and often outperforms the multi-task
oracle. This observation indicates
that Seq. FT does not degrade, and often enhances, the pretrained model’s generalization
capabilities.

\begin{figure}[t]
    \centering
    % ---- Left figure with its own caption ----
    \begin{subfigure}[t]{0.32\textwidth}
        \centering
        \includegraphics[width=\textwidth]{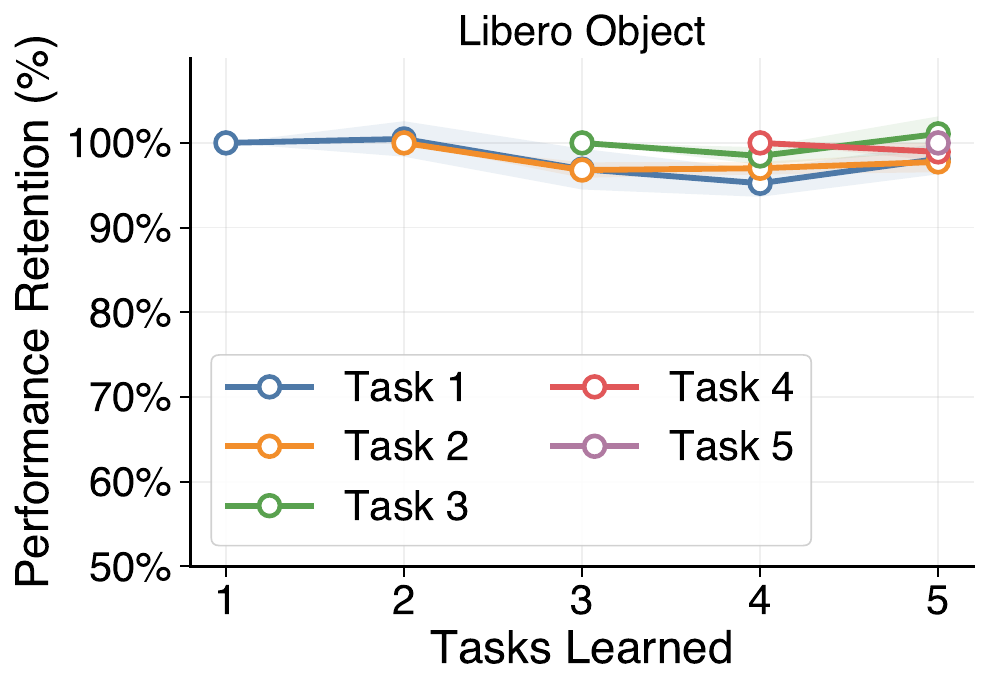}
    \end{subfigure}\hfill
    % ---- Left figure with its own caption ----
    \begin{subfigure}[t]{0.32\textwidth}
        \centering
        \includegraphics[width=\textwidth]{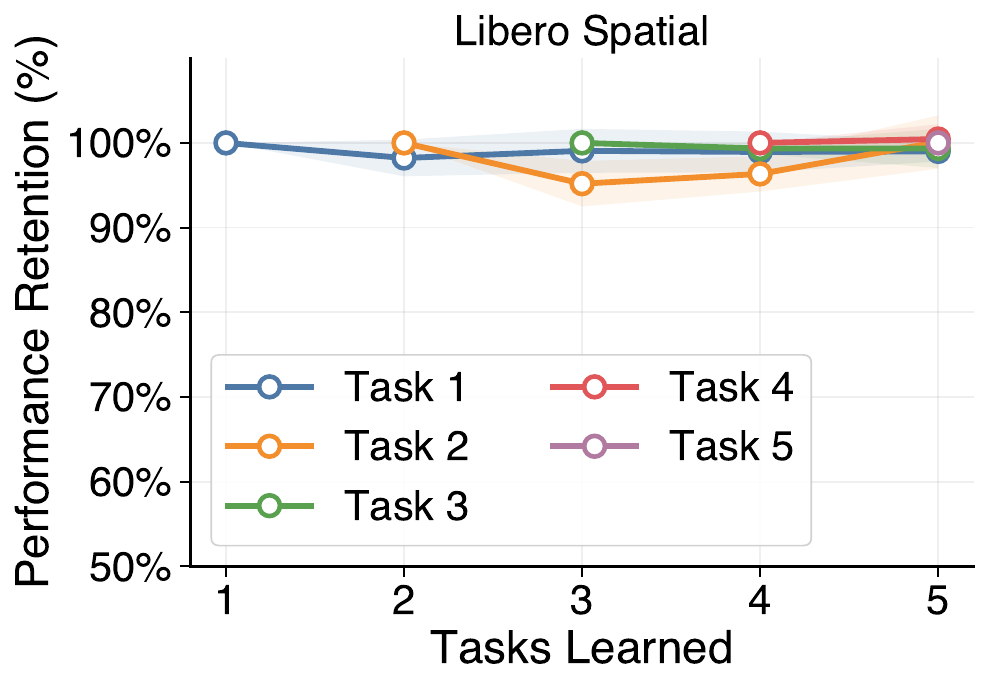}
    \end{subfigure}\hfill
    \begin{subfigure}[t]{0.32\textwidth}
        \centering
        \includegraphics[width=\textwidth]{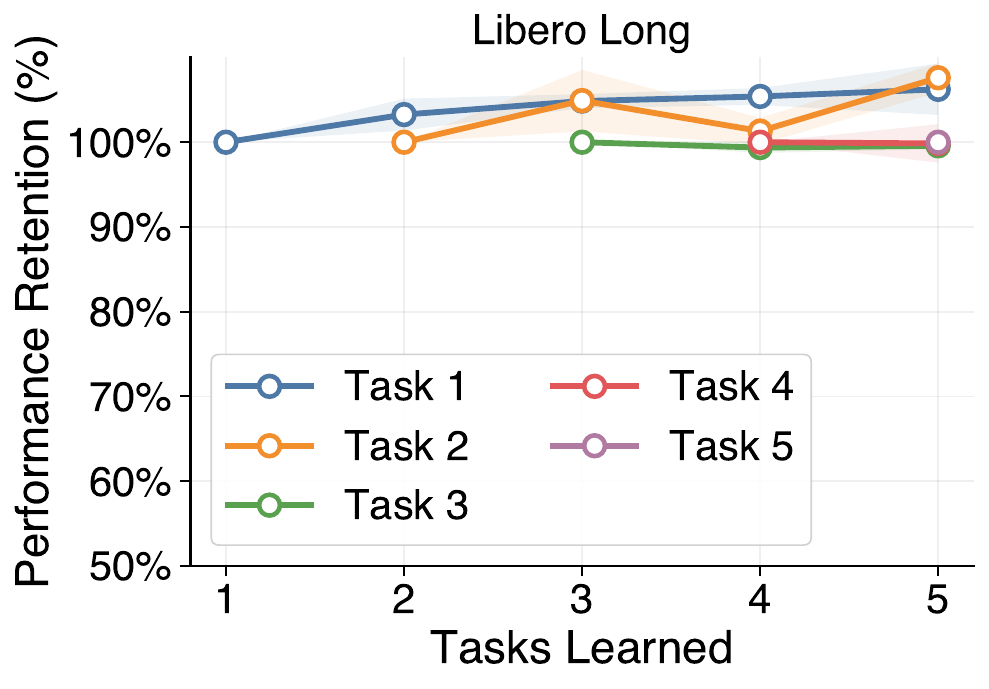}
    \end{subfigure}

    \caption{Each line tracks a single training task's success rate, normalized to 100$\%$ at the point it was first learned. Subsequent x-values show how that task's performance changes as additional tasks are learned. Sequential Fine-Tuning shows little forgetting throughout the entire training.}
    \label{fig:retention_plots}
    \vspace{-2em}
\end{figure}

Such surprisingly strong performance stems from the fact that naive Sequential Fine-Tuning exhibits almost \textbf{no forgetting} in these experiments (Fig.~\ref{fig:retention_plots}).
Contrary to the conventional expectation that Sequential
Fine-Tuning suffers from severe catastrophic forgetting,
we observe little performance degradation on previously
learned tasks, with the NBT metric consistently showing less than 2\% of (and sometimes even negative) forgetting. 
Given the absence of significant forgetting, it is therefore reasonable that Sequential Fine-Tuning performs competitively. Since it imposes no constraints or regularization on parameter updates, the optimization process can focus entirely on fitting the current task without incurring stability–plasticity trade-offs.

By contrast, \textbf{the addition of CRL techniques does not provide much added benefit} and often hurt the performance.
EWC, SLCA, and RETAIN all suffer a significant loss in plasticity, as illustrated by their lower average success rate due to constrained parameter updates. DWE cannot benefit from positive transfer due to parameter isolation. Replay-based methods require access to expert demonstrations and storage that grows with the number of tasks, yet do not improve performance. 
% Such a loss in plasticity is reasonable, as most of these CRL methods explicitly trade off plasticity for better retention of previously learned capabilities,  e.g., by imposing strong constraints on parameter updates.

Together, these results suggest that Sequential Fine-Tuning could be a strong
minimal-assumption approach for continual post-training of large
VLA models.
The observation that Sequential Fine-Tuning simultaneously exhibits
little forgetting, good plasticity, and preserved
generalization challenges conventional expectations in continual
learning. A natural question is whether this behavior is specific
to the three evaluated benchmarks, or whether it reflects a more
general property of large pretrained models trained with on-policy RL.

To examine the robustness of this phenomenon, we next introduce
a series of controlled variations to the training setup, including
environmental perturbations, changes of physical engine and VLA models, and task-order
modifications. As we will show, the favorable properties of Sequential
Fine-Tuning persist under these variations.
Finally, in Sec.~\ref{sec:analysis}, we provide mechanistic analysis
and additional empirical evidence to better understand the source
of this unexpected stability.

\subsection{Robustness Under Controlled Perturbations}
\label{ss:perturb}

To assess whether the strong performance of Sequential Fine-Tuning
depends on specific benchmark configurations, we conduct additional
experiments under controlled perturbations. We examine three axes of
variation: (1) environmental perturbations that alter visual and
state conditions, (2) changes in domain and model architecture,
and (3) modifications to the task order in the continual sequence.
Across all settings, we evaluate whether the three key properties
observed earlier, namely minimal forgetting, good plasticity,
and preserved zero-shot generalization, continue to hold.

\textbf{Environmental Perturbations.}
First, we assess the robustness of our result to changes in environment parameters across tasks. Specifically, we introduce three types of perturbation: \emph{camera perturbation}, where the camera position and orientation of each task is set to different values; \emph{lighting perturbation}, where the lighting intensity of each task is different; and \emph{robot state perturbation}, where the location of the robot base is different for each task.
These experiments evaluate whether the strong performance of Sequential Fine-Tuning is attributable to the environment parameters remaining constant in the original LIBERO benchmark.

\begin{table}[t]
\caption{Examining the consistency of Seq. FT performance across different perturbations. We \textbf{bold} the metrics for which Seq. FT outperforms the multitask oracle.}
\label{tab:perturb}
\centering
\renewcommand{\arraystretch}{1.3}
\footnotesize
\begin{tabular}{lcccccc}
\toprule

\textbf{Domain} / Method & \multicolumn{6}{c}{\textbf{Metrics (\%)}} \\
\cmidrule(lr){2-7}
  \quad & \textbf{AVG} $\uparrow$ & \textbf{$\Delta$AVG} $\uparrow$ & \textbf{NBT} $\downarrow$ & \textbf{FWT} $\uparrow$ & \textbf{ZS} $\uparrow$ & \textbf{$\Delta$ZS} $\uparrow$ \\
\midrule

\multicolumn{7}{l}{\textbf{Camera Perturbation}} \\
\quad Seq. FT & \textbf{75.5$\pm$0.2} & \textbf{+18.9} & -0.5$\pm$0.5 & 3.7$\pm$1.1 & \textbf{46.7$\pm$0.2} & \textbf{-0.6} \\
\rowcolor{gray!10} \quad Multitask (Oracle) & 75.2$\pm$0.1 & +18.6 & -- & -- & 43.8$\pm$0.5 & -3.6 \\

\midrule

\multicolumn{7}{l}{\textbf{Lighting Perturbation}} \\
\quad Seq. FT & 82.4$\pm$0.5 & +26.7 & 0.2$\pm$0.4 & 5.7$\pm$0.1 & \textbf{54.9$\pm$1.0} & \textbf{+1.9} \\
\rowcolor{gray!10} \quad Multitask (Oracle) & 87.0$\pm$0.3 & +31.3 & -- & -- & 54.1$\pm$0.3 & +1.2 \\

\midrule

\multicolumn{7}{l}{\textbf{Robot State Perturbation}} \\
\quad Seq. FT & 81.2$\pm$0.9 & +23.4 & 0.6$\pm$0.5 & 0.2$\pm$0.3 & \textbf{42.7$\pm$0.7} & \textbf{+2.4} \\
\rowcolor{gray!10} \quad Multitask (Oracle) & 86.1$\pm$0.3 & +28.3 & -- & -- & 42.2$\pm$0.7 & +1.9 \\

\midrule
\midrule

\multicolumn{7}{l}{\textbf{Pi-0 on RoboCasa}} \\
\quad Seq. FT & 29.5$\pm$3.0 & +10.6 & -0.1$\pm$2.1 & 1.2$\pm$1.7 & \textbf{21.5$\pm$1.9} & \textbf{+2.7} \\
\rowcolor{gray!10}\quad Multitask (Oracle) & 31.4$\pm$2.3 & +12.5 & -- & -- & 20.8$\pm$1.2 & +2.0 \\

\midrule

\multicolumn{7}{l}{\textbf{OpenVLA on ManiSkill}} \\
\quad Seq. FT & 70.9$\pm$1.5 & +19.4 & -1.0$\pm$1.5 & 0.5$\pm$0.6 & \textbf{51.0$\pm$0.8} & \textbf{+11.0} \\
\rowcolor{gray!10} \quad Multitask (Oracle) & 72.8$\pm$0.2 & +21.2 & - & - & 50.7$\pm$0.8 & +10.7 \\

\midrule
\midrule

\multicolumn{7}{l}{\textbf{Task Order Perturbation}} \\
\quad Seq. FT (Re-order 1) & 79.8$\pm$0.5 & +22.9 & 1.4$\pm$1.4 & 3.5$\pm$0.3 & \textbf{54.4$\pm$0.8} & \textbf{+3.9} \\
\quad Seq. FT (Re-order 2) & 81.2$\pm$1.0 & +24.4 & 1.6$\pm$1.7 & 0.8$\pm$1.3 & \textbf{55.7$\pm$0.5} & \textbf{+4.2} \\
\quad Seq. FT (Re-order 3) & 80.2$\pm$1.0 & +23.3 & -0.3$\pm$0.5 & 2.4$\pm$1.3 & \textbf{57.6$\pm$1.0} & \textbf{+6.1} \\
\rowcolor{gray!10} \quad Multitask (Oracle) & 85.8$\pm$0.2 & +28.9 & -- & -- & 51.2$\pm$0.7 & -0.3 \\

\bottomrule
\end{tabular}
\label{tab:success_rates}
\end{table}

\textbf{Domain and Model Variations.}
Next, we examine whether our conclusion still holds on different VLAs and in different benchmarks. In particular, besides the OpenVLA-OFT~\citep{kim2025finetuningvisionlanguageactionmodelsoptimizing} model that we used for experiments in Sec.~\ref{ss:main_rslt}, we additionally evaluate Pi-0~\citep{black2024pi_0}, a flow-matching VLA\footnote{We provide additional discussion about properties of flow-matching VLAs in Sec.~\ref{sss:diffusion}} built on PaliGemma, and OpenVLA~\citep{kim2024openvla}, an auto-regressive VLA based on Llama 2 that, unlike OpenVLA-OFT, does not use action chunking. We evaluate these models on the RoboCasa~\citep{nasiriany2024robocasa}, a benchmark with diverse scenes and many none-pick-and-place tasks, and Maniskill~\citep{gu2023maniskill2}, a benchmark based on the SAPIEN~\citep{xiang2020sapien} physical engine, respectively.

\textbf{Task Order Sensitivity.}
Finally, we investigate the sensitivity of Sequential Fine-Tuning
to task ordering. Classical continual learning methods often exhibit
strong dependence on the order in which tasks are presented,
particularly when tasks differ in difficulty or similarity. We construct alternative task sequences by permuting the order of
tasks within the libero-spatial benchmark and repeat the continual training
procedure.

We evaluate Sequential Fine-Tuning and the multi-task oracle under these perturbations, and report the results for these experiments in Table~\ref{tab:perturb}.
Across all conditions, Seq. FT maintains strong performance. 
Specifically, the \textbf{AVG} of Seq. FT consistently show a big increase from the base model, and maintains a $<5\%$ gap with the multi-task oracle (which, as discussed in Sec.~\ref{ss:gap}, can be bridged).
The \textbf{NBT} stays below $2\%$ for all experiments, with frequent negative values, indicating the same absence of catastrophic
forgetting that we noticed earlier.
Finally, the \textbf{ZS} performance maintains a consistent edge over the multitask oracle, demonstrating the surprising ability of Seq. FT to boost generalization.
Taken together, these robustness experiments indicate that the unexpected stability of Sequential Fine-Tuning is not a fragile artifact of benchmark design, but a \textbf{consistent pattern across environmental, architectural, and sequential variations}.

% We therefore turn to a mechanistic analysis to better understand the source of this behavior.

%% file: 4_2_add_exp.tex
\subsection{Additional Experiments and Discussions}
\label{ss:additional}

In this section, we present additional results and discussions regarding closing the training gap with oracle (Sec.~\ref{ss:gap}), the properties of VLAs with continuous diffusion head (Sec.~\ref{sss:diffusion}), and increasing the length of training task sequence (Sec.~\ref{sss:long}).

% Finally, we discuss how we can potentially close the already small gap on Average Success between the Sequential Fine-Tuning method and the Multitask Oracle.

\subsubsection{Closing the Training Gap Between the Multi-task Oracle and Sequential Fine-Tuning}
\label{ss:gap}

\begin{wrapfigure}{r}{0.5\textwidth}
\vspace{-1.0em}
\centering
% , trim=110 80 300 130, clip
\includegraphics[width=0.50\textwidth]{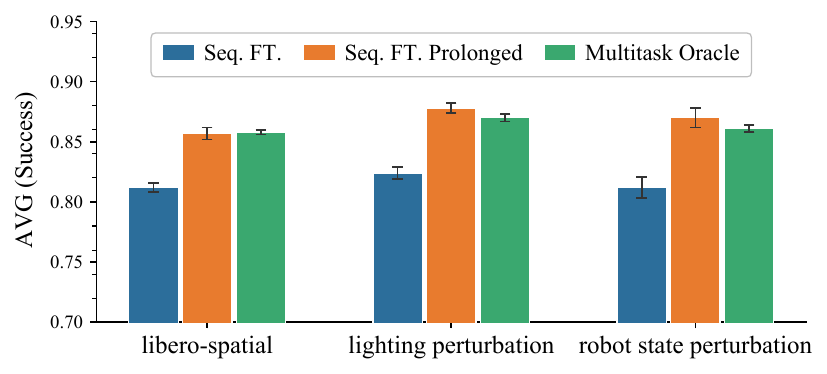}
\vspace{-1.0em}
\caption{Final training success rates: by simply prolonging the Seq. FT training steps, we can obtain on-par performance with multitask oracle. 
}
\label{fig:prolong}
% \vspace{-4.0em}
\end{wrapfigure}

In our experiments, we noted that there is a small but consistent gap between multi-task training and continual learning on the training task average success.
While it is understandable that CRL methods would under-perform the oracle, in this section we seek to investigate whether this gap is introduced by fundamental limitations of the CRL setup that caused the agent to converge to sub-optimal local optima. Specifically, we examine this question in the three domains where \textbf{the gap between the Seq. FT and the multitask oracle is largest} (around 5\%). We test whether we can bridge this gap by simply doubling the number of training episodes on the lowest performing task in each of these benchmarks, and report the results in Fig.~\ref{fig:prolong}.

% \begin{figure}[H]

As shown by these results, \textbf{we can close this gap and reach on-par AVG with the multitask oracle simply by training for more episodes}.
These results indicate that the AVG gap is not due to Seq. FT getting stuck at sub-optimal solutions. Instead, they highlight two insights: first, multi-task training may introduce synergies that improve sample efficiency, which is an intriguing direction for future study; second, if the goal is to match multi-task performance, Sequential Fine-Tuning can achieve it by simply training for more episodes on the lower-performing tasks.

\subsubsection{Properties of VLAs with Continuous Diffusion Heads}
\label{sss:diffusion}
In this work, we primarily consider VLA models with discrete auto-regressive action output that are structurally similar to LLMs and VLMs. While the same recipe can work for VLAs with diffusion head (as shown in Sec.~\ref{ss:perturb}), we empirically found that the continuous diffusion head often requires more careful constraints (e.g. using a lower LoRA rank). We conjecture that this sensitivity stems from the continuous denoising objective and high expressivity of diffusion-based action heads, which may make them more prone to policy drift under flexible adaptation. We leave a more systematic investigation of this effect to future work.

% We find VLAs with continuous diffusion heads don't always obey the same trends as autoregressive token-based VLAs under Seq. FT with PEFT. Notably, we observe that LoRA fine-tuning only the VLM backbone yields less catastrophic forgetting than full LoRA fine-tuning, which in turn yields less forgetting than LoRA fine-tuning only the diffusion action head. This suggests the diffusion action head is more susceptible to forgetting under sequential fine-tuning than the VLM backbone. We leave a full investigation of this phenomenon as an interesting direction for future work.

\subsubsection{Training on Longer Task Sequences}
\label{sss:long}
In this section, we examine whether our conclusions from the previous sections are robust to an extended continual learning horizon. We test whether Seq. FT can maintain little forgetting even when learning a large number of tasks through a 30 task scenario from libero-spatial, libero-object, and libero-long-horizon suites and present the results in Fig.~\ref{fig:long}. The results indicate that Seq. FT maintains high knowledge retention even as the number of sequential tasks scales to 30.

%% file: 5_analysis.tex
\section{Analysis: What Makes Sequential Fine-Tuning So Effective?}
\label{sec:analysis}
Given the experimental results in Sec.~\ref{sec:exp}, we conduct analysis and additional experiments in this section towards better understanding the surprising effectiveness of Sequential Fine-Tuning. We focus our analysis from the following three properties of Sequential Fine-Tuning in our experiments: little catastrophic forgetting, strong plasticity, and good zero-shot generalization. In the following sections, we discuss and analyze the reasons behind each of these properties.

% \begin{itemize}
%     \item Why is sequential finetuning robust to catastrophic forgetting? What contribute to this robustness?
%     \item Why is CRL still able to maintain good plasticity when constrained by LoRA?
%     \item Why is sequential finetuning able to achieve good zero-shot generalization?
% \end{itemize}

\subsection{Why Little Catastrophic Forgetting?}
\label{ss:forget}

\begin{wrapfigure}{r}{0.4\textwidth}
\vspace{-1.5em}
\centering
% , trim=110 80 300 130, clip
\includegraphics[width=0.40\textwidth]{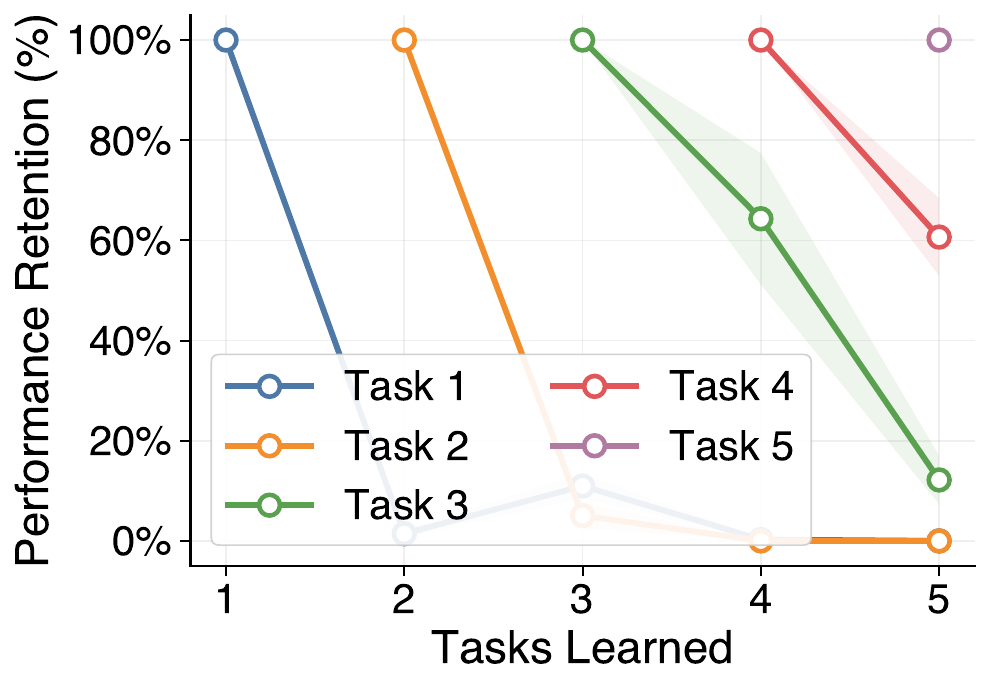}
\vspace{-2em}
\caption{VLA, on-policy RL, and LoRA are all crucial to stability. Removing any one of them results in catastrophic forgetting, shown by the curve.
}
\label{fig:forget}
\vspace{-1.0em}
\end{wrapfigure}

Most previous CRL methods are designed to mitigate catastrophic forgetting, with results showing that Sequential Fine-Tuning leads to significant unlearning of previous tasks. This mismatch raises a key question: why can simple Sequential Fine-Tuning avoid catastrophic forgetting in our experiments in the VLA domain? To investigate this phenomenon, we start by conducting ablation studies by (1) removing the RL objective (reducing to supervised fine-tuning), (2) replacing the large VLA model with a smaller neural network with 12 Million parameters, pre-trained to a similar (but inevitably slightly different) initial performance, and (3) removing LoRA. We describe the detailed setup of these experiments in Appendix~\ref{app:abla}, and show the results in Tab.~\ref{tab:forget}.

\begin{table}[H]
% \vspace{-1em}
\caption{Ablation studies on the libero-spatial benchmark}
\centering
\renewcommand{\arraystretch}{1.3}
\footnotesize
\begin{tabular}{lcccccc}
\toprule
\textbf{Ablation} & \multicolumn{6}{c}{\textbf{Metrics}} \\
\cmidrule(lr){2-7}
& \textbf{AVG} $\uparrow$ & \textbf{$\Delta$AVG} $\uparrow$ & \textbf{NBT} $\downarrow$ & \textbf{FWT} $\uparrow$ & \textbf{ZS} $\uparrow$ & \textbf{$\Delta$ZS} $\uparrow$ \\
\midrule
Seq. FT (Original) & \textbf{81.2±0.4} & \textbf{+24.3} & \textbf{0.3±0.5} & \textbf{0.3±0.5} & \textbf{57.1±1.1} &  \textbf{+5.6} \\

Supervised fine-tuning instead of RL & 29.9$\pm$2.3 & -27.0 & 78.7$\pm$1.9 & -53.8$\pm$0.0 & 1.1$\pm$0.9 & -50.4 \\

Smaller Policy & 13.1±0.9 & -53.7 & 11.4$\pm$3.7 & -63.4±0.5 & 0.0$\pm$0.0 & -56.2 \\

Without LoRA & 7.3$\pm$5.2 & -49.6 & 40.9$\pm$11.8 & -50.4$\pm$1.3 & 0.0$\pm$0.0 & -51.5 \\

\bottomrule
\end{tabular}
\label{tab:forget}
% \vspace{-1em}
\end{table}

The results here are revealing: all three components play a crucial role. Removing any one of them leads to a significant drop in both AVG performance and zero-shot generalization, where the model quickly loses all pre-trained capabilities during RL finetuning (Fig.~\ref{fig:forget}). In the following paragraphs, we analyze how each factor contributes to mitigating catastrophic forgetting.

% Experiment wise, we first show the performance after training on task 1. Then we show the final performance...

\textbf{Effect of On-Policy RL: }
The observation that on-policy RL helps prevent forgetting has been noted in several recent papers in the LLM domain~\citep{shenfeld2025rlsrazoronlinereinforcement, chen2025retaining, lai2026reinforcementfinetuningnaturallymitigates}. While no previous work has demonstrated this phenomenon in the VLA domain, it is perhaps not surprising that a similar conclusion holds.

% \bo{You can refer again to \citep{shenfeld2025rlsrazoronlinereinforcement}, so mention explicitly it is their and others claims, in case they are wrong hhh. Basically SFT is min $KL(\pi_b || \pi_\theta)$, while online RL is minimizing $KL(q || \pi_\theta)$, where $q \propto \pi_0 R$ (or $KL(\pi_\theta || q)$, $q \propto \pi_0 \exp(R)$ if we have an additional KL term as in GRPO). From this it is easy to see that SFT has no guarantee for distance between $\pi_\theta$ and $\pi_0$ (the initial policy from previous tasks, hence no stability in CL). While RL respect $\pi_0$, let alone modern RL algorithm adds clipping (trust-region update) and KL term explicity to further constrains update.
% }

As pointed out in \cite{shenfeld2025rlsrazoronlinereinforcement}, this effect can largely be attributed to the use of on-policy data.
Specifically, let $\pi_0(a\mid s)$ denote the base policy and $\pi_\theta(a\mid s)$ the adapted policy. Supervised fine-tuning learns with
\[
\nabla_\theta \mathcal{L}_{\mathrm{SFT}}
=
- \mathbb{E}_{(s,a)\sim D_{\text{task}}}
\big[
\nabla_\theta \log \pi_\theta(a \mid s)
\big].
\]
Thus, supervised fine-tuning increases the log-probability of dataset actions regardless of how small $\pi_0(a\mid s)$ was. 
If the dataset contains actions outside the high-probability region of $\pi_0$, probability mass must be shifted into regions where $\pi_0(a\mid s)$ is small. 
This necessarily increases the forward KL divergence:
\[
\mathrm{KL}\!\left(\pi_\theta \,\|\, \pi_0\right)
=
\mathbb{E}_{s}
\;
\mathbb{E}_{a\sim\pi_\theta(\cdot\mid s)}
\left[
\log \frac{\pi_\theta(a\mid s)}{\pi_0(a\mid s)}
\right],
\]
which grows when $\pi_\theta$ allocates mass to actions unlikely under $\pi_0$.

The policy gradient update, by contrast, results in
\[
\nabla_\theta J(\theta)
=
\mathbb{E}_{s\sim d_{\pi_\theta},\, a\sim\pi_\theta}
\big[
A^{\pi_\theta}(s,a)\,
\nabla_\theta \log \pi_\theta(a\mid s)
\big].
\]

where $d_{\pi_\theta}$ is the on-policy state distribution and 
$A^{\pi_\theta}(s,a)$ is the advantage function.
Crucially, both the objective and its gradient are weighted by samples 
$(s,a) \sim d_{\pi_\theta}(s)\pi_\theta(a \mid s)$. 
In other words, policy gradient updates only reweight probability mass where $\pi_\theta$ already has support, and cannot suddenly assign high probability to actions with near-zero probability. As a result, the probability mass can only move gradually outward from the support of $\pi_0$, creating an implicit objective that minimizes KL drift from $\pi_0$. Since forgetting empirically correlates with forward KL from $\pi_0$~\citep{shenfeld2025rlsrazoronlinereinforcement}, such an implicit regularization helps the model retain its learning capability and mitigate catastrophic forgetting.

While it is impressive that RL helps alleviate catastrophic forgetting, it is equally worth noticing that, unlike in previous work~\citep{shenfeld2025rlsrazoronlinereinforcement,chen2025retaining,lai2026reinforcementfinetuningnaturallymitigates}, our results on the VLA domains suggest that ~\emph{on-policy RL alone is not sufficient for avoiding catastrophic forgetting}, and both the large pretrained model and parameter-efficient adaptation (i.e., LoRA) are also critical for maintaining performance.

\textbf{Effect of Large Pretrained Models: }
The effect of large pretrained models for mitigating forgetting can be largely attributed to the curse (or rather “blessing” in our case) of dimensionality~\citep{mirzadeh2022wide}. Specifically, for two random unit vectors $u, v \in \text{Unif}(\mathcal{S}^{d-1})$, it is well known that $\sqrt{d}\langle u, v \rangle 
\rightarrow \mathcal{N}\!\left(0, 1\right)$ as $d \rightarrow \infty$. In other words, in high-dimensional space, almost all random vectors are nearly orthogonal. As a result, overparametrized models inherently create a vast “Null Space” where gradient updates in most directions barely affect the pre-trained knowledge, as also noted in concurrent work~\citep{liu2026pretrainedvisionlanguageactionmodelssurprisingly}. 

We empirically validate this analysis via examining the Fisher Information~\citep{kirkpatrick2017overcoming}.
Let $\boldsymbol{\theta} \in \mathbb{R}^D$ denote the model parameters, $\mathbf{g} = \nabla_{\boldsymbol{\theta}} \mathcal{L}(\boldsymbol{\theta})$ the gradient of the loss of the \textit{current training task}, and $\mathbf{F} \in \mathbb{R}^{D \times D}$ denote the Fisher Information Matrix (FIM) with respect to the \textit{pre-training tasks}. Using a local second-order approximation, the increase in the pre-training loss under a parameter update $\Delta$ can be written as
\[
L_{\text{old}}(\boldsymbol{\theta}+\Delta)
\approx
L_{\text{old}}(\boldsymbol{\theta})
+ \tfrac{1}{2}\Delta^\top \mathbf{F}\Delta .
\]
Thus, if the current task updates parameters along direction $\mathbf{g}$, the resulting increase in the old-task loss is governed by $\mathbf{g}^\top \mathbf{F}\mathbf{g}$. We therefore compute the Rayleigh quotient of the Fisher Information Matrix along the gradient direction as

\[
E_F(\mathbf{g})
=
\frac{\mathbf{g}^\top \mathbf{F} \mathbf{g}}{\mathbf{g}^\top \mathbf{g}}
=
\frac{\sum_{d=1}^D f_d \, g_d^2}{\sum_{d=1}^D g_d^2}.
\]

We define $E_F(\mathbf{g})$ as the 
 \emph{Fisher energy}, which measures the average curvature of the pre-training tasks along the gradient direction of the current task, and therefore quantifies how strongly the new task will interfere with the pretrained knowledge, where a high value indicates more interference. 

Since the full FIM scales quadratically with the number of parameters, we use a diagonal empirical approximation for the FIM:
$
\mathbf{F} \approx \mathrm{diag}(f_1, \dots, f_D), 
\quad \text{where} \quad 
f_d = \mathbb{E}\!\left[g_d^2\right],
$
and normalize it by $\max_d(f_d)$ so that the value is in $[0, 1]$. 
We examine $E_F(\mathbf{g})$ for both the small neural network policy from ablation study, and the large OpenVLA-OFT model on the libero-spatial task suite.
On the large OpenVLA-OFT model, the average $E_F$ is only 0.02, indicating very little interference between the task gradient and pretrained knowledge. However, on the small policy, $E_F$ jumps to 0.16, which likely explains the catastrophic forgetting that occurs with small models.

\textbf{Effect of Low-Rank Adaptation:}
LoRA constrains fine-tuning updates to a low-rank subspace, restricting the gradient update $\Delta W$ to a rank-$r$ subspace around the pretrained weight $W_0$. By concentrating task-specific changes within this narrow, low-dimensional subspace ($r \ll d$), LoRA limits the degrees of freedom of the update, preventing simultaneous alterations of the high-energy principal directions of the model. 
Therefore, it is perhaps not very surprising that LoRA can alleviate catastrophic forgetting and preserve pre-trained knowledge. However, our empirical analysis suggests that the effect of LoRA may be deeper than this simple interpretation. Rather than merely reducing the effective rank of the update, LoRA appears to \emph{prevent a small subset of layers from undergoing disproportionately large structural changes} during fine-tuning.

We examine this hypothesis by empirically analyzing the weight update $\Delta W$ obtained with different LoRA rank, and without using LoRA (i.e. full fine-tuning). We show the results in Tab.~\ref{tab:lora_study}.

%%%%%%%%% Need to think about how to present this
\begin{table}[H]
% \vspace{-1em}
\caption{Examining the properties of the delta weight with and without LoRA}
\centering
\renewcommand{\arraystretch}{1.0}
\footnotesize
\begin{tabular}{l|cccc}
\toprule
\scriptsize \textbf{Method} & \scriptsize\textbf{Effective Rank (mean)} & \scriptsize\textbf{Effective Rank (std)} & \scriptsize\textbf{Nuclear Norm} & \scriptsize\textbf{NBT} $\downarrow$ \\

\midrule
LoRA Rank 32 (default) & 27.5 & 5.7 & 0.48 & 0.3 \\
Full Finetuning & 324.7 & 465.0 & 4.31 & 40.9 \\
LoRA Rank 512 & 303.4 & 89.3 & 0.42 & 0.6 \\
\bottomrule
\end{tabular}
\label{tab:lora_study}
% \vspace{-1em}
\end{table}

We observe that LoRA (with rank 512) achieves substantially better NBT performance than full fine-tuning
(0.6\% vs 40.9\%), even though their mean per-layer effective ranks are comparable
(303.4 vs. 324.7).
A closer inspection reveals an important difference: the across-layer standard deviation
of effective rank is much larger for full fine-tuning than for LoRA (465.0 vs. 89.3).
This result suggests that full fine-tuning produces highly uneven update geometry across the
network. In particular, a subset of layers undergoes extremely high-rank updates, which
may correspond to substantial structural modification of the pretrained representations
in those layers. Such uneven, high-rank updates likely result in the overwriting of previously
learned knowledge.

By contrast, LoRA maintains a much lower across-layer standard deviation, thereby constraining the per-layer update geometry and preventing any individual layer from undergoing uncontrolled high-rank structural modification. Consistent with this preserved geometry, LoRA also yields a lower nuclear norm, indicating a smaller total magnitude of directional modification per layer, and potentially lead to the preservation of previously acquired knowledge.

To summarize, \textbf{RL, LoRA, and the VLA itself alleviate catastrophic forgetting from three complementary perspectives: objective, constraints, and capacity}. As a result, their synergistic combination leads to stable learning without forgetting in a way that no two of them alone exhibit, as we empirically observe in our experiments.

\subsection{Why Good Plasticity?}
The ability of Sequential Fine-Tuning to learn new tasks effectively is well-known \citep{liu2023libero}, but it is more surprising that this good plasticity is preserved even when LoRA is applied. In particular, previous studies have noted that “LoRA often underperforms in supervised pre-training” \citep{biderman2024lora}, where the constrained gradient update reduces the plasticity of the model. This contrast raises the question of why our model, with LoRA applied, is still able to learn effectively and maintain high plasticity in continual Reinforcement Learning.

Upon further investigation, we found that such a result is tightly coupled with the nature of policy gradient RL, and more specifically to its low-capacity requirements. We follow \citet{schulman2025lora}, and illustrate this phenomenon from an information-theoretic perspective. Specifically, policy gradient methods such as GRPO learn based on the advantage
function, which only provides O(1) bits of information for each episode under a sparse reward setup. For example, in our experiments on OpenVLA-OFT with 7B parameters, the rank-32 LoRA weights contain around 100M parameters, which is more than enough to absorb the information obtained from the 50k training rollout episodes.
By contrast, in supervised learning, the information contained in each episode scales linearly with the length of the episode, and therefore often leads to per-episode information that is thousands of time richer than in RL. Such a discrepancy likely leads to the performance loss of LoRA when applied to supervised learning in previous work. This perspective highlights the synergy between on-policy RL and LoRA, as their combination effectively reduces catastrophic forgetting without sacrificing much plasticity.

\subsection{Why Good Zero-shot Generalization?}

Finally, we observe that Sequential Fine-Tuning consistently preserves strong zero-shot generalization. Since maintaining zero-shot capability can be viewed as a form of preventing forgetting, this behavior can largely be understood through the same mechanisms discussed in Sec.~\ref{ss:forget}.
What is more intriguing is that Sequential Fine-Tuning often maintains a slight edge over oracle multi-task training on the generalization capabilities. Although this gap is generally small on the benchmarks we evaluate, the trend is consistent across settings and therefore noteworthy. We do not yet have a definitive explanation for this phenomenon. One plausible hypothesis is that task sequencing acts as a form of implicit regularization. Rather than jointly optimizing over all tasks and potentially overfitting to the aggregated objective, sequential training exposes the model to a shifting objective over time~\citep{abel2023definition}. Such non-stationary optimization dynamics may encourage more robust representations and improved generalization. Investigating this implicit regularization effect more rigorously remains an exciting direction for future work.

%% file: 6_appendix.tex
\appendix

\begin{figure}[h]
% \vspace{-1em}
\centering
\includegraphics[width=1.0\textwidth]{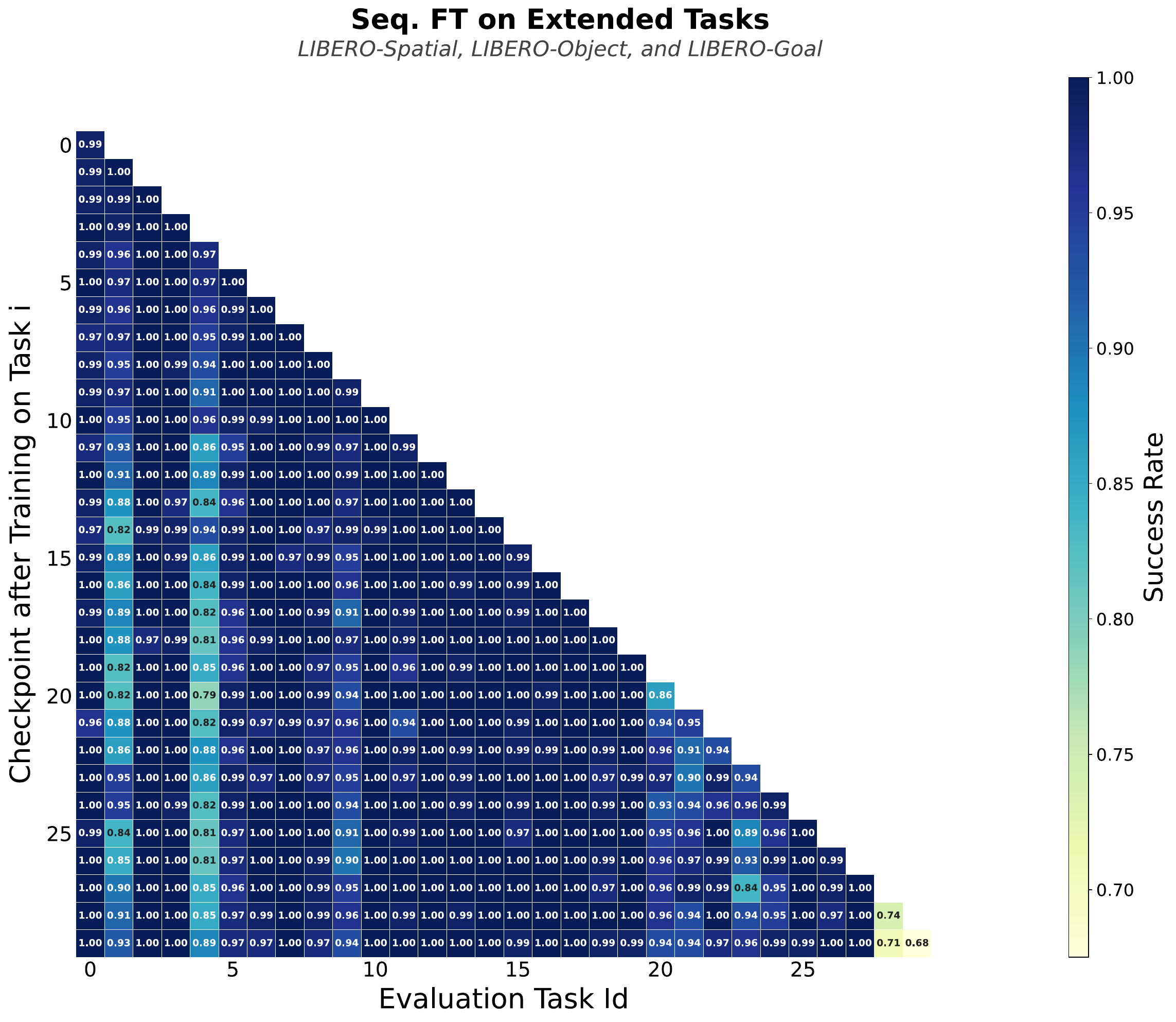}
\caption{
Performance on tasks after sequentially training on 30 tasks. Seq. FT remains robust under extended length of the training task sequence.
}
\label{fig:long}
% \vspace{-2em}
\end{figure}

\section{GRPO Training Formulation}
\label{app:grpo}
In GRPO, at each update, trajectories are sampled from the previous policy $\pi_{\theta_{\text{old}}}$, and the policy is optimized using
\[
\max_\theta \;
\mathbb{E}_{(s_t,a_t)\sim \pi_{\theta_{\text{old}}}}
\left[
\min\!\left(
\rho_t(\theta)\,\hat{A},
\;\mathrm{clip}(\rho_t(\theta),1-\epsilon,1+\epsilon)\,\hat{A}
\right)
\right],
\]
where
\[
\rho_t(\theta)
=
\frac{\pi_\theta(a_t \mid s_t,\ell)}
{\pi_{\theta_{\text{old}}}(a_t \mid s_t,\ell)},
\quad
\hat{A}
=
\frac{R - \mu_R}{\sigma_R}.
\]
Here $R$ denotes the episodic return of the sampled trajectory, and $\mu_R,\sigma_R$ are the mean and standard deviation of returns within the sampled group.

For VLA models that generate actions via autoregressive tokens~\citep{kim2024openvla,kim2025finetuningvisionlanguageactionmodelsoptimizing}, GRPO can be applied directly by treating the sequence of action tokens as the policy output and computing the likelihood ratios over tokens. For VLA models with continuous flow or diffusion action heads~\citep{black2024pi_0,intelligence2025pi}, actions are generated by integrating a learned velocity field defined by a deterministic ordinary differential equation (ODE):
\[
\frac{dx_t}{dt} = v_\theta(x_t, t).
\]
Since deterministic flows do not provide stochastic exploration required by policy gradients, we adopt the Flow-SDE formulation~\citep{chen2026pitextttrlonlinerlfinetuning} and introduce controlled Gaussian noise into the dynamics:
\[
dx_t = v_\theta(x_t, t)\,dt + \sigma_t\, dW_t,
\]
where $\sigma_t$ is a noise schedule and $dW_t$ is a Wiener process increment.
This converts the deterministic sampler into a stochastic policy that defines a distribution over actions.
Standard policy gradient objectives (e.g., PPO or GRPO) can then be applied by optimizing the advantage-weighted likelihood over the resulting action trajectories.

% \section{Retention Visualization}
% \label{app:ret_vis}

% \begin{figure}[H]
%     \centering
%     % ---- Left figure with its own caption ----
%     \begin{subfigure}[t]{0.32\textwidth}
%         \centering
%         \includegraphics[width=\textwidth]{images/retention_plots/libero_object_Naive_LoRA_retention_bands.pdf}
%     \end{subfigure}\hfill
%     % ---- Left figure with its own caption ----
%     \begin{subfigure}[t]{0.32\textwidth}
%         \centering
%         \includegraphics[width=\textwidth]{images/retention_plots/libero_spatial_Naive_LoRA_retention_bands.pdf}
%     \end{subfigure}\hfill
%     \begin{subfigure}[t]{0.32\textwidth}
%         \centering
%         \includegraphics[width=\textwidth]{images/retention_plots/libero_long_Naive_LoRA_retention_bands.pdf}
%     \end{subfigure}

%     \caption{Each line tracks a single training task's success rate, normalized to 100$\%$ at the point it was first learned (its peak post-training performance). Subsequent x-values show how that task's performance changes as additional tasks are learned. A flat line at 100$\%$ indicates no forgetting. Sequential Fine-Tuning shows little forgetting throughout the entire training.}
%     \label{fig:retention_plots}
% \end{figure}

\section{Evaluation Metrics}
\label{app:ev_metric}

Suppose tasks arrive sequentially in the order 
$\{\mathcal{T}_1, \dots, \mathcal{T}_T\}$. 
After completing training on task $\mathcal{T}_i$, we evaluate the policy 
on all tasks $\mathcal{T}_j$ and record the success rate 
$S_{i,j} \in [0,1]$. 
This produces a success matrix $S \in \mathbb{R}^{T\times T}$, 
where $S_{i,j}$ denotes the success rate on task $j$
after training up to task $i$. Additionally, we denote the initial performance of the base model on task $j$ as $S_{0,j}$.

\paragraph{Training Average Final Success (AVG).}
The overall performance after learning all tasks is defined as the 
average final success rate:
\begin{equation}
\text{AVG} = \frac{1}{T} \sum_{j=1}^{T} S_{T,j}.
\end{equation}
This measures how well the final policy performs across the entire training task sequence.

\paragraph{Negative Backward Transfer (NBT).}
Negative Backward Transfer (a.k.a Forgetting) measures the degradation in performance on previous tasks after learning subsequent ones. 
Since each task is trained once in sequence, 
we define forgetting relative to the performance immediately 
after completing training on that task:
\begin{equation}
\text{NBT} = 
\frac{1}{T-1} \sum_{j=1}^{T-1}
\left( S_{j,j} - S_{T,j} \right).
\end{equation}
Lower values indicate better retention of previously acquired skills, where a value of 0 indicate that there is no forgetting on average.

\paragraph{Forward Transfer (FWT).}
Forward transfer quantifies whether learning previous tasks improves 
performance on future tasks before they are trained. 
Let $S_{0,j}$ denote the zero-shot success rate on task $j$ 
before any task-specific training. Then
\begin{equation}
\text{FWT} = \frac{1}{T-1} \sum_{j=2}^{T} 
\left( S_{j-1,j} - S_{0,j} \right).
\end{equation}
Positive values indicate beneficial transfer to unseen tasks. Importantly, FWT is strongly influenced by the task ordering. To better measure transfer capabilities, we propose an additional metric called the held-out performance, as explained below.

\paragraph{Held-Out Tasks Performance (ZS).}
Unlike in classic continual RL, VLA contain strong zero-shot performance on unseen tasks even before any training occur.
To evaluate the ability to retain and potentially enhance these zero-shot capabilities, 
we assess the final policy on a set of held-out tasks 
$\mathcal{H}$ not encountered during continual training. 
Held-out performance is defined as
\begin{equation}
\text{ZS} = 
\frac{1}{|\mathcal{H}|} 
\sum_{h \in \mathcal{H}} S^{\text{held}}_{T,h},
\end{equation}
where $S^{\text{held}}_{T,h}$ denotes the success rate on held-out task $h$
after completing training on all tasks. 

%%%%%%%%%%%%%%%%%%%%%%%%%%%%%%%%%%%%%%%%%%%%%%%%%%%%%%%%%%
\section{Evaluation Algorithms}
\label{app:algo}
In this section, we describe the algorithms we evaluated in our study in detail, as well as the reasoning for choosing these algorithms. We begin by establishing two reference points that anchor our evaluation.

\textbf{Sequential Fine-Tuning: }
The most direct approach to continual learning is to train tasks
sequentially without any additional mechanism to prevent forgetting.
At each stage, the model is fine-tuned solely on the current task via interaction. Sequential Fine-Tuning requires no replay buffer, parameter isolation, or task-specific regularization.
It is commonly treated as a lower-bound baseline in continual RL,
as it is expected to suffer from catastrophic forgetting under
non-stationary task sequences.

\textbf{Multi-Task Training (Oracle):}
As an upper-bound reference, we train a model jointly on all tasks,
assuming simultaneous access to experiences from the entire task set.
This setting violates the sequential and non-stationary assumptions
of continual learning and therefore serves as an oracle baseline.
Its performance is often used to represent the best achievable performance when task order constraints are removed.

Beyond these reference points, we evaluate a diverse set of continual
learning algorithms spanning the principal methodological paradigms
in the literature. Continual reinforcement learning (CRL) methods are commonly categorized
into three principal paradigms~\citep{pan2025survey}: (i) \emph{regularization-based methods},
which constrain parameter updates to preserve prior knowledge;
(ii) \emph{replay-based methods}, which reuse data or model outputs from
previous tasks; and (iii) \emph{parameter-isolation methods}, which
allocate task-specific capacity to avoid interference.
To systematically evaluate these paradigms, we evaluate the
following representative approaches.

\begin{itemize}
    \item \textbf{Elastic Weight Consolidation~\citep{kirkpatrick2017overcoming}} (regularization-based): 
    penalizes parameter updates directly in the weight space, in proportion to their estimated importance to previous tasks using a Fisher-based quadratic constraint.

    \item \textbf{Expert Replay~\citep{rolnick2019experience}} (replay-based): stores expert demonstrations for all tasks and replay them during training as an additional Behavior Cloning loss term. Note that this approach requires access to the expert demonstrations, as well as space to store the demonstration data which grows linearly with the number of tasks.

    \item \textbf{Dark Experience Replay~\citep{buzzega2020dark}} (replay-based):
    Instead of replaying labels, DER matches the logits of the previous
    model, preserving functional behavior while avoiding the use of expert data.
    Note that this approach requires storing previous interactions and logit values which grow linearly with the number of tasks.

    \item \textbf{Dynamic Weight Expansion} (parameter isolation):
     We allocate an isolated task-specific LoRA adapter~\citep{hu2021loralowrankadaptationlarge} for each task that is only activated when facing the corresponding task, thereby preventing interference in gradient updates. The number of adapter weights grows linearly with the number of tasks.
    
\end{itemize}

In addition to classical CRL methods, we evaluate two additional methods
motivated by recent advances in large pretrained models:

\begin{itemize}
    \item \textbf{SLCA~\citep{zhang2023slca}}:
    a method for layerwise learning-rate decoupling, by applying higher learning rates to action head and lower
    rates to the VLM trunk, in an effort to preserve the pre-trained representations of the base VLA model.

    \item \textbf{RETAIN~\citep{yadav2025robust}}:
    after training on each task, RETAIN merges the delta weight update back into the base model with a discount coefficient, instead of fully accepting it.
    RETAIN represents model-merging approaches designed to balance
    adaptation and retention in weight space without explicit replay
    or importance estimation.
\end{itemize}

Together, these methods span the dominant CRL paradigms as well as
emerging large-model adaptation strategies.

%%%%%%%%%%%%%%%%%%%%%%%%%%%%%%%%%%%%%%%%%%%%%%%%%%%%%%%%%%%

\section{Experiment Setup}
\label{app:setup}

Each of our base models are obtained by performing supervised fine-tuning with a small amount of in-domain data, so that the model has non-zero initial success rate. This setup allows us to examine performance across a range of initial policy qualities and verify that our results are not specific to a single checkpoint. Here we provide the detailed experiment setup across different benchmarks.

\begin{table}[h]
\centering
\footnotesize
\caption{Experiment setup across benchmarks.}
\begin{tabular}{l|ccccc}
\toprule
Parameter & libero-object & libero-spatial & libero-long-horizon & RoboCasa & maniskill \\
\midrule
Base Model & OpenVLA-OFT & OpenVLA-OFT & OpenVLA-OFT & Pi-0 & OpenVLA \\
\# of SFT Demos & 10 & 10 & 432 & 240 & 140 \\
Initial Training Success & 55.6 & 56.9 & 83.0 & 18.9 & 51.6 \\
\# of Training Tasks & 5 & 5 & 5 & 4 & 4 \\
Episodes per Task & 10240 & 10240 & 5120 & 3840 & 10240 \\
Episode Length & 512 & 512 & 512 & 480 & 80 \\
\bottomrule
\end{tabular}
\label{tab:evaluation_setup}
\end{table}

For ManiSkill, we standardize the plate, background, and table color and restrict the variation in initial states to 40 discrete object positions and 4 object rotations. This reduces evaluation variance and ensures that all methods are evaluated on the same fixed set of task configurations, improving the comparability and reproducibility of results. We opt to use 4 training tasks which allows us to run multiple seeds and baselines while keeping the total experimental budget tractable.

We select training tasks whose initial success rates are neither near zero nor saturated. Tasks with non-zero initial performance ensure that the base model already possesses some relevant capabilities, allowing CRL to refine existing behaviors rather than learning entirely from scratch. Avoiding tasks with near-saturated performance leaves sufficient headroom for improvement, making it possible to meaningfully evaluate learning throughout training.

\section{Ablation Setup}
\label{app:abla}
We provide additional details for the ablation experiments used in the libero-spatial benchmark. Unless otherwise specified, all ablations use the same environment setup, evaluation protocol, and shared hyperparameters described in Appendix~\ref{app:shared_hype}.

\begin{table}[h]
\centering
\footnotesize
\caption{Ablation setup for libero-spatial benchmark.}
\begin{tabular}{l|ccc}
\toprule
Parameter & Supervised fine-tuning instead of RL & Smaller Policy & Without LoRA \\
\midrule
Base Model & 7B OpenVLA-OFT & 12M CNN with MLP Head & 7B OpenVLA-OFT \\
\# of Training Tasks & 5 & 5 & 5 \\
Pre-training Demos & 10 & 30 & 10 \\
Initial Training Success & 56.9 & 66.8 & 56.9 \\
Batch Size & 256 & 8192 & 8192 \\
RL Episodes per Task & - & 10240 & 10240 \\
\midrule
SFT Dataset Demos & 432 & - & - \\
SFT Training Steps & 600 & - & - \\
\bottomrule
\end{tabular}
\label{tab:ablation_setup}
\end{table}

\textbf{Supervised fine-tuning instead of RL} replaces online RL post-training with supervised fine-tuning on a dataset of demonstration trajectories collected from the environment. The policy is fine-tuned via behavior cloning on 432 demonstration trajectories using the same base model and input representation as the RL setup.

\textbf{Smaller Policy} replaces the OpenVLA-OFT model with a small CNN policy of around 12M parameters. The policy is initially supervised finetuned on 30 demonstrations to prime the model with non-zero success and RL finetuned using the same setup as Seq. FT.

\textbf{Without LoRA} performs RL post-training on the full OpenVLA model without parameter-efficient LoRA adapters, instead updating the base model parameters directly. All other RL hyperparameters remain identical to the main experimental setup.

\section{Shared Hyperparameter}
\label{app:shared_hype}
Here we present hyperparameters for the shared components of VLA post-training. These settings are used across all tasks unless otherwise specified. We adopt GRPO as the base algorithm and LoRA adapters with rank 32. Other hyperparameters follow the standard configuration listed below.

\begin{table}[H]
\centering
\caption{Hyperparameters for RL post-training.}
\begin{small}
\begin{tabular}{ccc}
\toprule
\textbf{Algorithm} & \textbf{Name} & \textbf{Value} \\
\midrule
\multirow{15}{*}{GRPO}
 & Optimizer & AdamW \\
 & Learning rate & $2\times10^{-5}$ \\
 & AdamW $\beta_1$ & 0.9 \\
 & AdamW $\beta_2$ & 0.999 \\
 & Adamw $\epsilon$ & $10^{-5}$ \\
 & Gradient clip norm & 1.0 \\
 & Global batch size & 8192 \\
 & Discount $\gamma$ & 0.99 \\
 & GAE $\lambda$ & 0.95 \\
 & Clip ratio (low/high) & 0.20 / 0.28 \\
 & KL coefficient $\beta$ & 0.0 \\
 & Entropy bonus & 0.0 \\
 & Rollout epochs & 16 \\
 & Group size & 8 \\
 & LoRA rank & 32 \\
\bottomrule
\end{tabular}
\end{small}
\label{tab:rl_hyperparameters}
\end{table}

\section{Method Hyperparameters}
\label{app:sepc_hype}
This table summarizes the method-specific hyperparameters used for each continual learning algorithm in our experiments. Sequential Fine-Tuning, Dynamic Weight Expansion, and multitask training are omitted, as they do not introduce any additional hyperparameters beyond those shared across all experiments.

% ~\citep{zheng2026revisiting}

\begin{table}[H]
\centering
\caption{Algorithm-specific hyperparameters.}
\begin{small}
\begin{tabular}{cccccc}
\toprule
\textbf{Algorithm} & \multicolumn{2}{c}{\textbf{Name}} & \multicolumn{2}{c}{\textbf{Value}} \\
\midrule

\multirow{2}{*}{EWC}
& \multicolumn{2}{c}{regularization coefficient $\lambda$} & \multicolumn{2}{c}{$1\times10^{6}$} \\
& \multicolumn{2}{c}{fisher estimation samples} & \multicolumn{2}{c}{65536} \\

\midrule

\multirow{3}{*}{ER}
& \multicolumn{2}{c}{Replay loss weight $\lambda_{\text{replay}}$} & \multicolumn{2}{c}{0.03} \\
& \multicolumn{2}{c}{Replay \# trajectories} & \multicolumn{2}{c}{10} \\
& \multicolumn{2}{c}{Replay global batch size} & \multicolumn{2}{c}{8192} \\
\midrule

\multirow{3}{*}{DER}
& \multicolumn{2}{c}{Replay loss weight $\lambda_{\text{replay}}$} & \multicolumn{2}{c}{0.03} \\
& \multicolumn{2}{c}{Replay \# trajectories} & \multicolumn{2}{c}{10} \\
& \multicolumn{2}{c}{Replay global batch size} & \multicolumn{2}{c}{8192} \\
\midrule

\multirow{2}{*}{SLCA}
& \multicolumn{2}{c}{slow learning rate} & \multicolumn{2}{c}{$4\times10^{-6}$} \\
& \multicolumn{2}{c}{fast learning rate} & \multicolumn{2}{c}{$4\times10^{-5}$} \\

\midrule

\multirow{1}{*}{RETAIN}
& \multicolumn{2}{c}{merge coefficient $\lambda$} & \multicolumn{2}{c}{0.5} \\

\bottomrule
\end{tabular}
\end{small}
\label{tab:algorithm_hyperparameters}
\end{table}

\section{Environment Description}
\label{app:env_desc}
In this section, we describe the environments used in our experiments, including task visualizations, natural language instructions, and the corresponding train-test splits.

\subsection{Libero-Spatial}
\subsubsection{Training Tasks}
\begin{enumerate}
    \taskwithimage {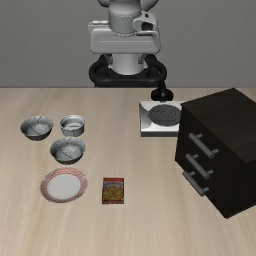}{pick up the black bowl between the plate and the ramekin and place it on the plate}
    \taskwithimage {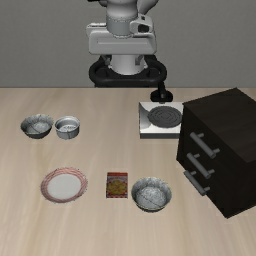}{pick up the black bowl next to the ramekin and place it on the plate}
    \taskwithimage {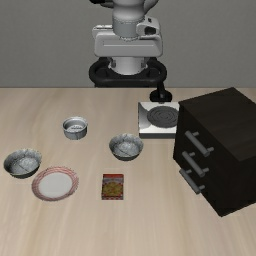}{pick up the black bowl from table center and place it on the plate}
    \taskwithimage {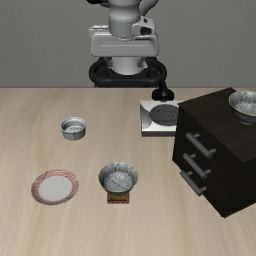}{pick up the black bowl on the cookie box and place it on the plate}
    \taskwithimage {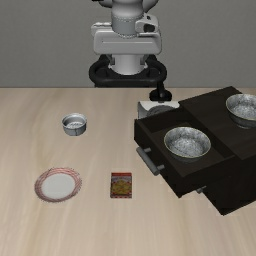}{pick up the black bowl in the top drawer of the wooden cabinet and place it on the plate}
\end{enumerate}
\subsubsection{Held-Out Tasks}
\begin{enumerate}
    \taskwithimage {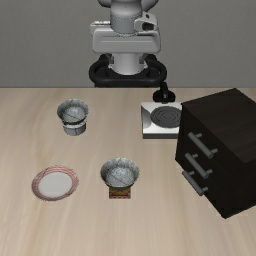}{pick up the black bowl on the ramekin and place it on the plate}
    \taskwithimage {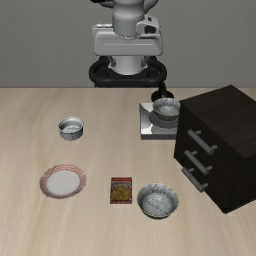}{pick up the black bowl next to the cookie box and place it on the plate}
    \taskwithimage {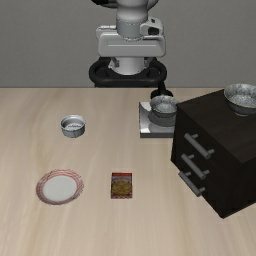}{pick up the black bowl on the stove and place it on the plate}
    \taskwithimage {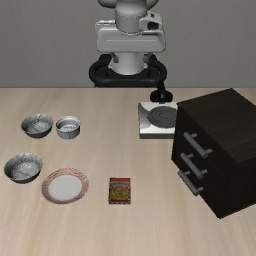}{pick up the black bowl next to the plate and place it on the plate}
    \taskwithimage {images/libero_spatial/pick_up_the_black_bowl_on_the_wooden_cabinet_and_place_it_on_the_plate.png}{pick up the black bowl on the wooden cabinet and place it on the plate}
\end{enumerate}

\subsection{Libero-Long}
\subsubsection{Training Tasks}
\begin{enumerate}
    \taskwithimage {images/libero_long/put_the_black_bowl_in_the_bottom_drawer_of_the_cabinet_and_close_it.png}{put the black bowl in the bottom drawer of the cabinet and close it}
    \taskwithimage {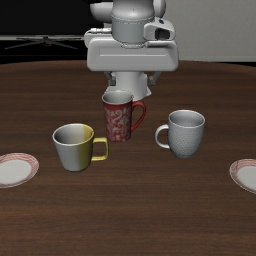}{put the white mug on the left plate and put the yellow and white mug on the right plate}
    \taskwithimage {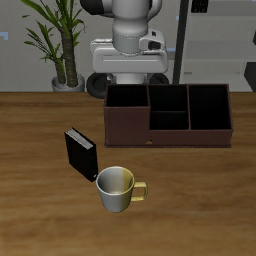}{pick up the book and place it in the back compartment of the caddy}
    \taskwithimage {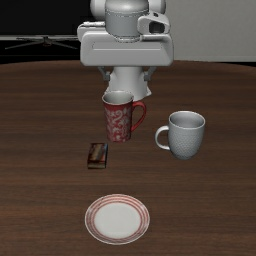}{put the white mug on the plate and put the chocolate pudding to the right of the plate}
    \taskwithimage {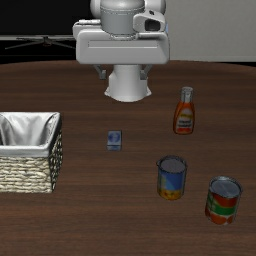}{put both the alphabet soup and the cream cheese box in the basket}
\end{enumerate}
\subsubsection{Held-Out Tasks}
\begin{enumerate}
    \taskwithimage {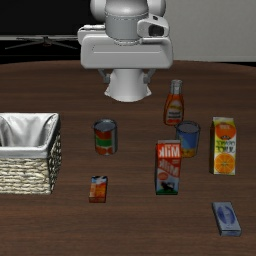}{put both the alphabet soup and the tomato sauce in the basket}
    \taskwithimage {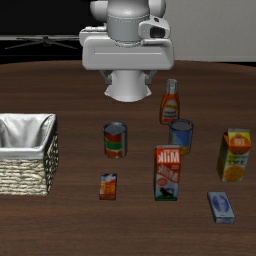}{put both the cream cheese box and the butter in the basket}
    \taskwithimage {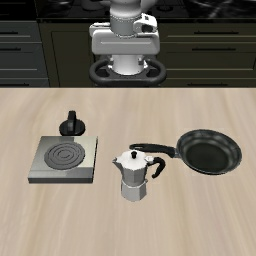}{turn on the stove and put the moka pot on it}
    \taskwithimage {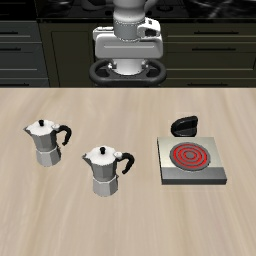}{put both moka pots on the stove}
    \taskwithimage {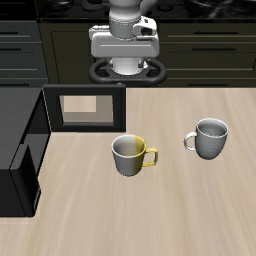}{put the yellow and white mug in the microwave and close it}
\end{enumerate}

\subsection{Libero-Object}
\subsubsection{Training Tasks}
\begin{enumerate}
    \taskwithimage {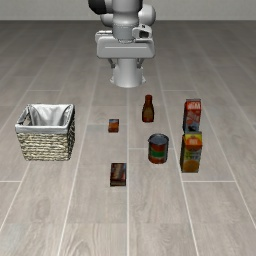}{pick up the tomato sauce and place it in the basket}
    \taskwithimage {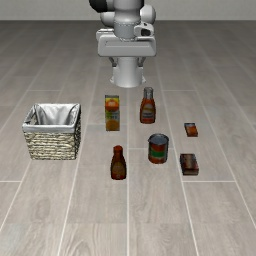}{pick up the butter and place it in the basket}
    \taskwithimage {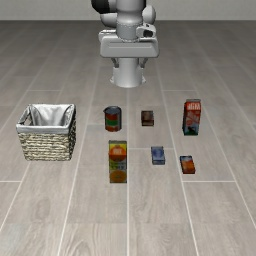}{pick up the milk and place it in the basket}
    \taskwithimage {images/libero_object/pick_up_the_chocolate_pudding_and_place_it_in_the_basket.png}{pick up the chocolate pudding and place it in the basket}
    \taskwithimage {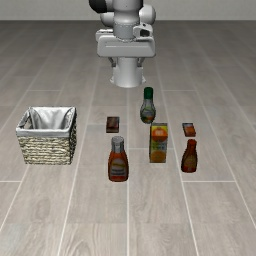}{pick up the orange juice and place it in the basket}
\end{enumerate}
\subsubsection{Held-Out Tasks}
\begin{enumerate}
    \taskwithimage {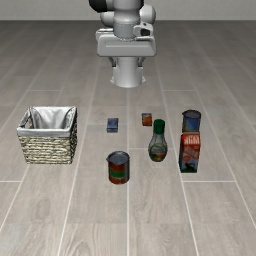}{pick up the alphabet soup and place it in the basket}
    \taskwithimage {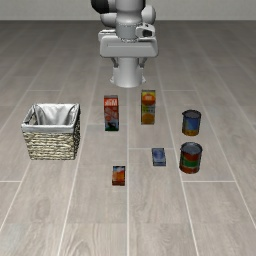}{pick up the cream cheese and place it in the basket}
    \taskwithimage {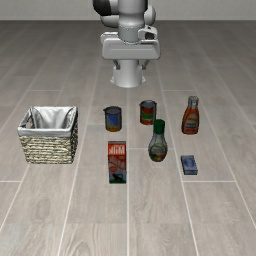}{pick up the salad dressing and place it in the basket}
    \taskwithimage {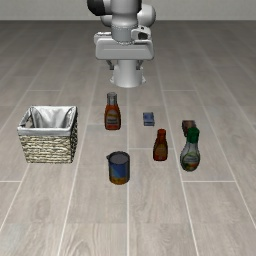}{pick up the bbq sauce and place it in the basket}
    \taskwithimage {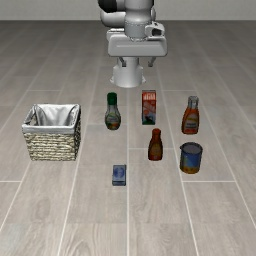}{pick up the ketchup and place it in the basket}
\end{enumerate}

\subsection{RoboCasa}
\subsubsection{Training Tasks}
\begin{enumerate}
    \taskwithimage {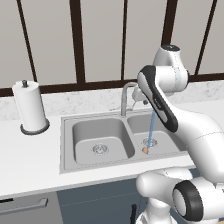}{turn sink spout}
    \taskwithimage {images/robocasa_task_viz/TurnSinkSpout.png}{turn on sink faucet}
    \taskwithimage {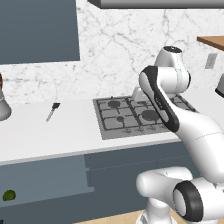}{close drawer}
    \taskwithimage {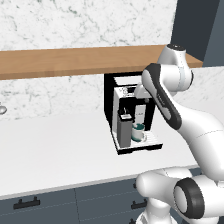}{press coffee machine button}
\end{enumerate}
\subsubsection{Held-Out Tasks}
\begin{enumerate}
    \taskwithimage {images/robocasa_task_viz/CloseSingleDoor.png}{close cabinet or microwave door}
    \taskwithimage {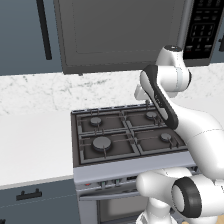}{turn on microwave}
    \taskwithimage {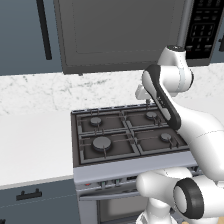}{turn off microwave}
    \taskwithimage {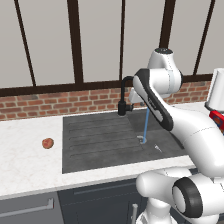}{turn off sink faucet}
\end{enumerate}

\subsection{Maniskill Put Plate On Scene 25 Main}
\subsubsection{Training Tasks}
\begin{enumerate}
    \taskwithimage {images/maniskill/carrot.png}{put carrot on plate}
    \taskwithimage {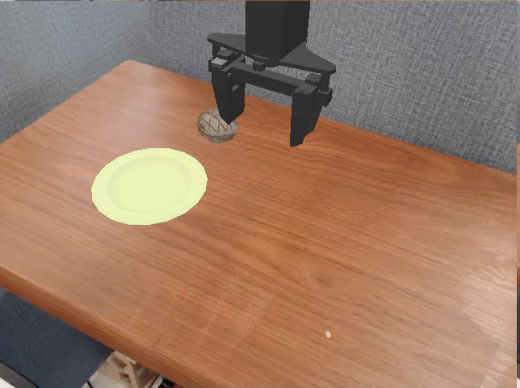}{put bread on plate}
    \taskwithimage {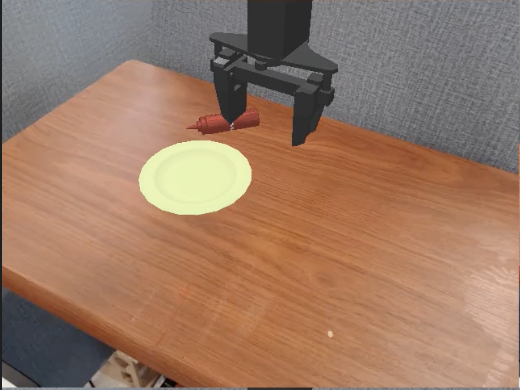}{put ketchup bottle on plate}
    \taskwithimage {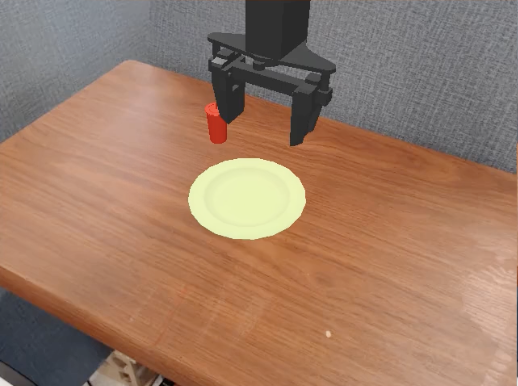}{put fast food cup on plate}
\end{enumerate}
\subsubsection{Held-Out Tasks}
\begin{enumerate}
    \taskwithimage {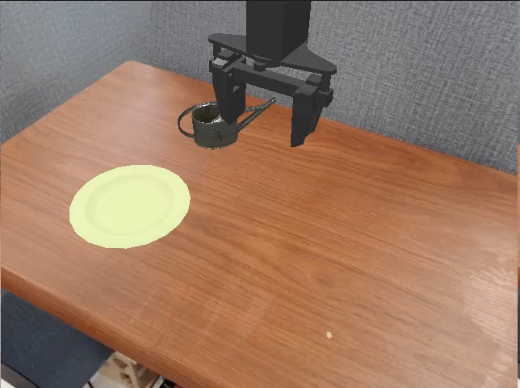}{put watering can on plate}
    \taskwithimage {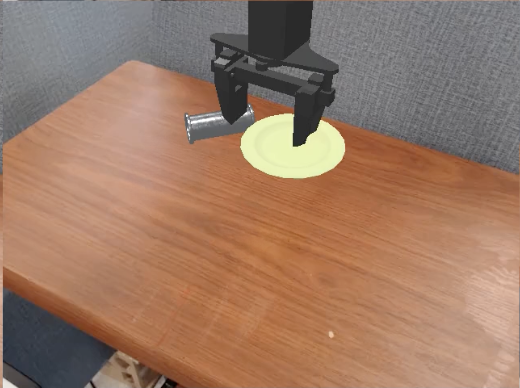}{put pipe on plate}
    \taskwithimage {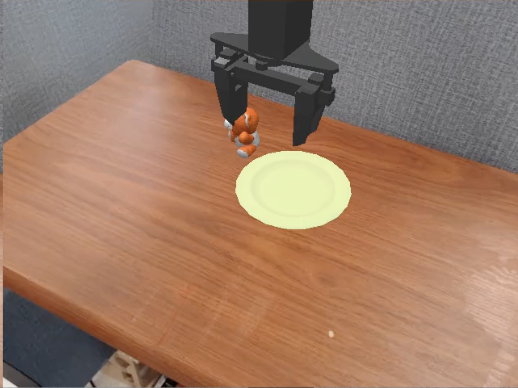}{put toy bear on plate}
    \taskwithimage {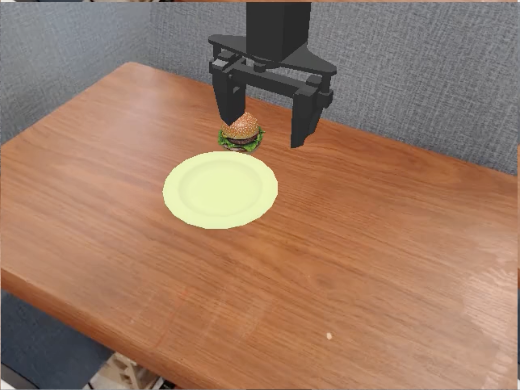}{put hamburger on plate}
\end{enumerate}

\subsection{Perturb Camera Angle}

\begin{figure}[H]
    \centering
    \begin{subfigure}[t]{0.13\textwidth}
        \centering
        \includegraphics[width=\textwidth]{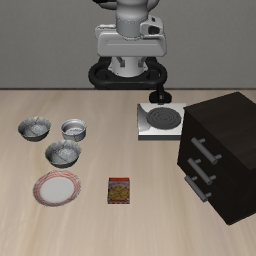}
    \end{subfigure}
    \begin{subfigure}[t]{0.13\textwidth}
        \centering
        \includegraphics[width=\textwidth]{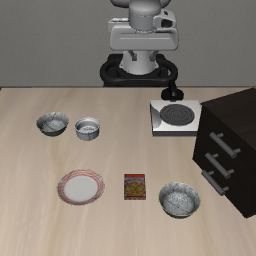}
    \end{subfigure}
    \begin{subfigure}[t]{0.13\textwidth}
        \centering
        \includegraphics[width=\textwidth]{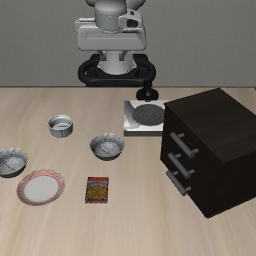}
    \end{subfigure}
    \begin{subfigure}[t]{0.13\textwidth}
        \centering
        \includegraphics[width=\textwidth]{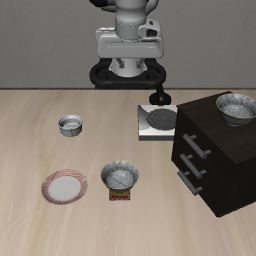}
    \end{subfigure}
    \begin{subfigure}[t]{0.13\textwidth}
        \centering
        \includegraphics[width=\textwidth]{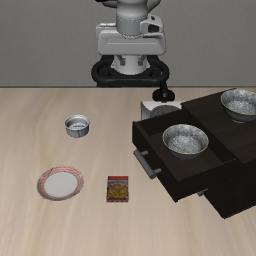}
    \end{subfigure}
    \\
    \begin{subfigure}[t]{0.13\textwidth}
        \centering
        \includegraphics[width=\textwidth]{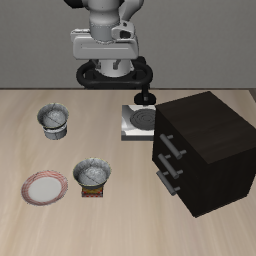}
    \end{subfigure}
    \begin{subfigure}[t]{0.13\textwidth}
        \centering
        \includegraphics[width=\textwidth]{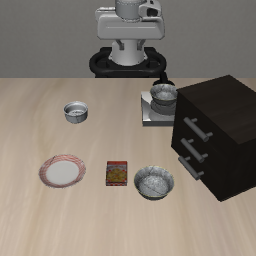}
    \end{subfigure}
    \begin{subfigure}[t]{0.13\textwidth}
        \centering
        \includegraphics[width=\textwidth]{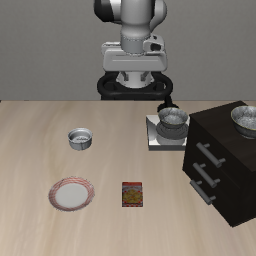}
    \end{subfigure}
    \begin{subfigure}[t]{0.13\textwidth}
        \centering
        \includegraphics[width=\textwidth]{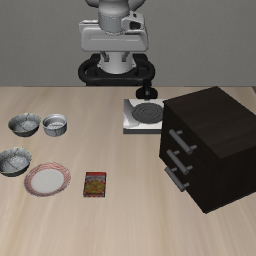}
    \end{subfigure}
    \begin{subfigure}[t]{0.13\textwidth}
        \centering
        \includegraphics[width=\textwidth]{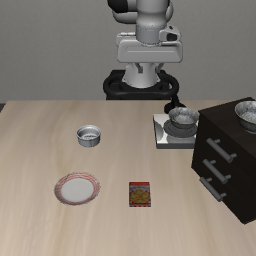}
    \end{subfigure}

    \caption{Changing camera angles.}
\end{figure}

\subsection{Perturb Lighting Conditions}

\begin{figure}[H]
    \centering
    \begin{subfigure}[t]{0.13\textwidth}
        \centering
        \includegraphics[width=\textwidth]{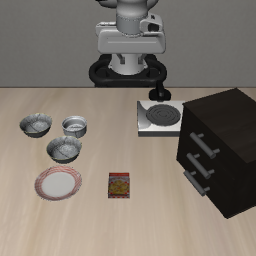}
    \end{subfigure}
    \begin{subfigure}[t]{0.13\textwidth}
        \centering
        \includegraphics[width=\textwidth]{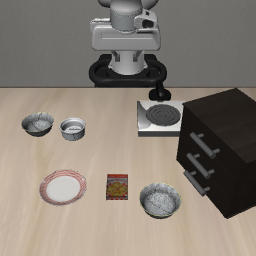}
    \end{subfigure}
    \begin{subfigure}[t]{0.13\textwidth}
        \centering
        \includegraphics[width=\textwidth]{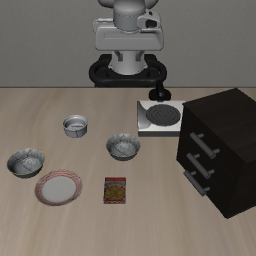}
    \end{subfigure}
    \begin{subfigure}[t]{0.13\textwidth}
        \centering
        \includegraphics[width=\textwidth]{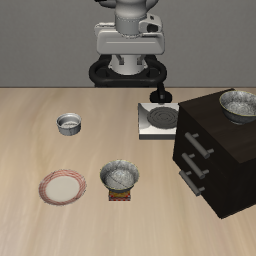}
    \end{subfigure}
    \begin{subfigure}[t]{0.13\textwidth}
        \centering
        \includegraphics[width=\textwidth]{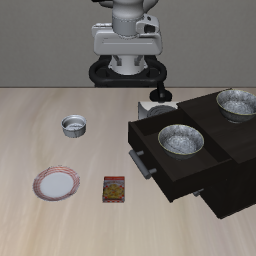}
    \end{subfigure}
    \\
    \begin{subfigure}[t]{0.13\textwidth}
        \centering
        \includegraphics[width=\textwidth]{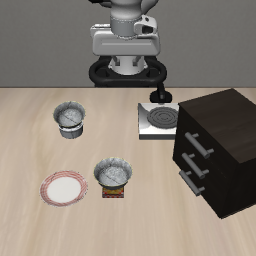}
    \end{subfigure}
    \begin{subfigure}[t]{0.13\textwidth}
        \centering
        \includegraphics[width=\textwidth]{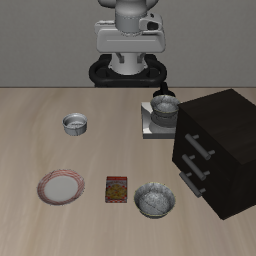}
    \end{subfigure}
    \begin{subfigure}[t]{0.13\textwidth}
        \centering
        \includegraphics[width=\textwidth]{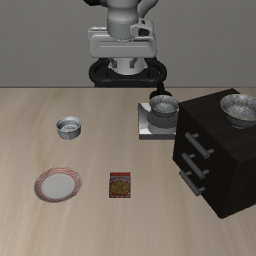}
    \end{subfigure}
    \begin{subfigure}[t]{0.13\textwidth}
        \centering
        \includegraphics[width=\textwidth]{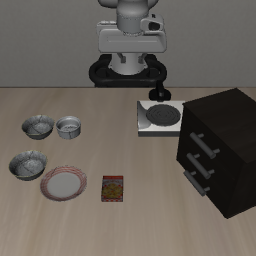}
    \end{subfigure}
    \begin{subfigure}[t]{0.13\textwidth}
        \centering
        \includegraphics[width=\textwidth]{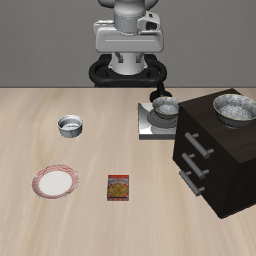}
    \end{subfigure}

    \caption{Changing Lighting conditions.}
\end{figure}

\subsection{Perturb Robot Position}

\begin{figure}[H]
    \centering
    \begin{subfigure}[t]{0.13\textwidth}
        \centering
        \includegraphics[width=\textwidth]{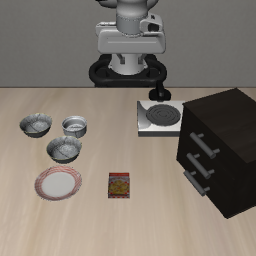}
    \end{subfigure}
    \begin{subfigure}[t]{0.13\textwidth}
        \centering
        \includegraphics[width=\textwidth]{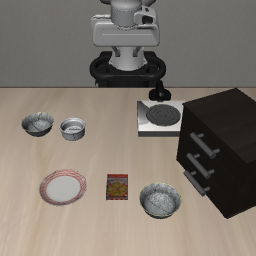}
    \end{subfigure}
    \begin{subfigure}[t]{0.13\textwidth}
        \centering
        \includegraphics[width=\textwidth]{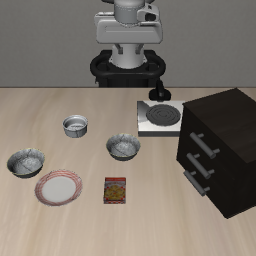}
    \end{subfigure}
    \begin{subfigure}[t]{0.13\textwidth}
        \centering
        \includegraphics[width=\textwidth]{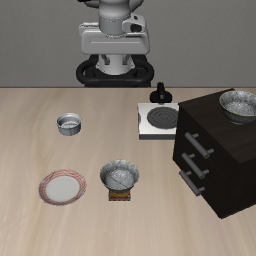}
    \end{subfigure}
    \begin{subfigure}[t]{0.13\textwidth}
        \centering
        \includegraphics[width=\textwidth]{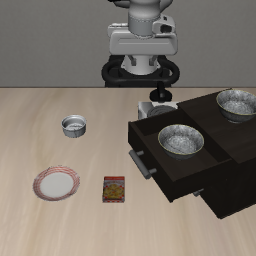}
    \end{subfigure}
    \\
    \begin{subfigure}[t]{0.13\textwidth}
        \centering
        \includegraphics[width=\textwidth]{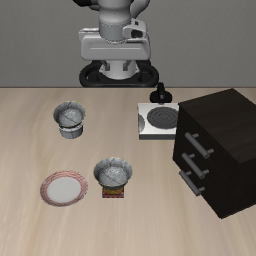}
    \end{subfigure}
    \begin{subfigure}[t]{0.13\textwidth}
        \centering
        \includegraphics[width=\textwidth]{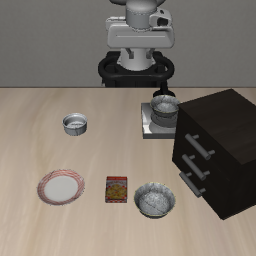}
    \end{subfigure}
    \begin{subfigure}[t]{0.13\textwidth}
        \centering
        \includegraphics[width=\textwidth]{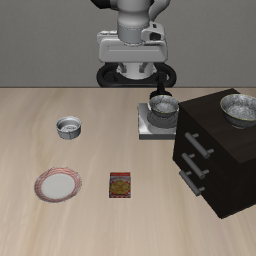}
    \end{subfigure}
    \begin{subfigure}[t]{0.13\textwidth}
        \centering
        \includegraphics[width=\textwidth]{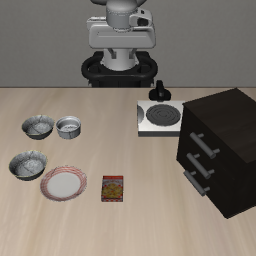}
    \end{subfigure}
    \begin{subfigure}[t]{0.13\textwidth}
        \centering
        \includegraphics[width=\textwidth]{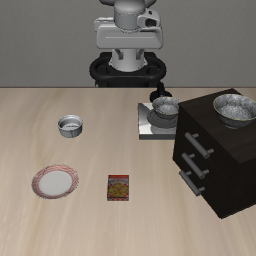}
    \end{subfigure}

    \caption{Changing Robot initial position.}
\end{figure}